\documentclass[11pt, a4paper, logo, copyright, nonumbering]{style}
\usepackage[authoryear, sort&compress, round]{natbib}
\usepackage{graphicx}
\usepackage{xspace}
\usepackage{xcolor}
\usepackage{adjustbox}
\usepackage{array}
\usepackage{float}
\usepackage{geometry}
\usepackage{dblfloatfix}
\usepackage{enumitem}
\usepackage{setspace}
\usepackage{footmisc}
\usepackage{booktabs}
\usepackage{multirow}
\usepackage{longtable}
\usepackage{tabularx}
\usepackage{xltabular}
\usepackage{threeparttable}
\usepackage{siunitx}

\usepackage{amsmath}
\usepackage{amssymb}
\usepackage{amsfonts}
\usepackage{mathtools}
\usepackage{bm}
\usepackage{dsfont}

\usepackage{subcaption}
\usepackage{tikz}
\usepackage{pgfplots}
\pgfplotsset{compat=1.18}
\usepackage{algorithm}
\usepackage{algpseudocode}
\usepackage{listings}

\usepackage{cleveref}
\usepackage{hyperref}
\hypersetup{colorlinks=true}

\usepackage{todonotes}
\usepackage{fontawesome5}
\usepackage[most]{tcolorbox}
\usepackage{CJKutf8}
\usepackage{lipsum}
\newcolumntype{L}[1]{>{\raggedright\let\newline\\\arraybackslash\hspace{0pt}}m{#1}}
\newcolumntype{R}[1]{>{\raggedleft\let\newline\\\arraybackslash\hspace{0pt}}m{#1}}

\newcommand{\ignore}[1]{}

\makeatletter
\DeclareRobustCommand\onedot{\futurelet\@let@token\@onedot}
\def\@onedot{\ifx\@let@token.\else.\null\fi\xspace}

\makeatother

\definecolor{MyBlue}{rgb}{0.46, 0.50, 0.61}
\definecolor{MyDarkBlue}{rgb}{0,0.08,0.8}
\definecolor{MyDarkGreen}{RGB}{45,155,45}
\definecolor{MyDarkRed}{rgb}{0.8,0.02,0.02}
\definecolor{MyOrange}{rgb}{1.0, 0.4, 0.2}
\definecolor{MyPurple}{RGB}{111,0,255}
\definecolor{MyRed}{rgb}{0.8,0.0,0.0}
\definecolor{MyGold}{rgb}{0.75,0.6,0.12}
\definecolor{MyDarkgray}{rgb}{0.66, 0.66, 0.66}
\definecolor{MyBrown}{rgb}{0.65, 0.16, 0.16}
\definecolor{MyMutedRose}{rgb}{0.58, 0.29, 0.35}
\definecolor{JiayuanColor}{rgb}{0.60,0.43,0.48}
\definecolor{erranColor}{rgb}{24, 40, 113}

\definecolor{citecolor}{HTML}{696FAD}
\newif\ifpropositionfirstitem
\propositionfirstitemtrue

\definecolor{bggray}{HTML}{F5F5F5}
\definecolor{pvdblue}{HTML}{DAE8FC}
\definecolor{RoseQuartzBg}{HTML}{F7CAC9}
\definecolor{RoseQuartz}{HTML}{F5A798}
\definecolor{Serenity}{HTML}{92A8D1}
\definecolor{OrangeRed}{rgb}{1.0, 0.27, 0.0}
\definecolor{RoyalBlue}{cmyk}{1, 0.50, 0, 0}
\definecolor{Turquoise}{HTML}{0F4C81}
\definecolor{mint}{rgb}{0.24, 0.71, 0.54}
\definecolor{green}{rgb}{0.0, 0.120, 0.0}

\newdimen\abovecrulesep
\newdimen\belowcrulesep
\makeatletter
\patchcmd{\@@@cmidrule}{\aboverulesep}{\abovecrulesep}{}{}
\patchcmd{\@xcmidrule}{\belowrulesep}{\belowcrulesep}{}{}
\makeatother

\definecolor{mybluetitle}{HTML}{4B527E} %{535C91} %{2C4E80}%{2A4C65} %{60656C} %{75809C}

\definecolor{codegreen}{HTML}{478058}%{rgb}{0,0.6,0}
\definecolor{codegray}{rgb}{0.5,0.5,0.5}
\definecolor{codepurple}{HTML}{4F5E80} %637A9F}%{rgb}{0.58,0,0.82}
\definecolor{backcolour}{rgb}{0.95,0.95,0.95}
\lstdefinestyle{mystyle}{
    backgroundcolor=\color{backcolour},
    commentstyle=\color{codegreen},
    keywordstyle=\color{magenta},
    numberstyle=\tiny\color{codegray},
    stringstyle=\color{codepurple},
    basicstyle=\ttfamily\scriptsize,
    breakatwhitespace=false,
    breaklines=true,
    captionpos=b,
    keepspaces=true,
    frame=none,
    numbersep=5pt,
    showspaces=false,
    showstringspaces=false,
    showtabs=false,
    tabsize=2
}
\newtcolorbox{promptbox}[2][]{
    enhanced, 
    breakable,
    center title,
    left*=0pt, right*=0pt,
    boxsep=2pt, left=5pt, right=5pt,
    skin first=enhanced,
    skin middle=enhanced,
    skin last=enhanced,
    colback  = backcolour,
    fonttitle=\bfseries\rmfamily,
    fontupper=\scriptsize,
    title={\footnotesize\strut{#2}},
    #1
    }
\newtcolorbox{onebox}[2][]{
    enhanced, 
    center title,
    left*=0pt, right*=0pt,
    boxsep=2pt, left=5pt, right=5pt,
    skin first=enhanced,
    skin middle=enhanced,
    skin last=enhanced,
    colframe = mybluetitle!90,
  colback  = mybluetitle!10,
    fonttitle=\bfseries\rmfamily\fontfamily{phv}\selectfont,
    title={\strut{#2}  \refstepcounter{subsubsection} \addcontentsline{toc}{subsubsection}{\string\numberline{\thesubsubsection}#2}
    },
    #1
    }

\makeatletter
\def\@BTrule[#1]{%
  \ifx\longtable\undefined
    \let\@BTswitch\@BTnormal
  \else\ifx\hline\LT@hline
    \nobreak
    \let\@BTswitch\@BLTrule
  \else
     \let\@BTswitch\@BTnormal
  \fi\fi
  \global\@thisrulewidth=#1\relax
  \ifnum\@thisruleclass=\tw@\vskip\@aboverulesep\else
  \ifnum\@lastruleclass=\z@\vskip\@aboverulesep\else
  \ifnum\@lastruleclass=\@ne\vskip\doublerulesep\fi\fi\fi
  \@BTswitch}
\makeatother

\addto\extrasenglish{
}

 {\begin{list}{}%
         {\setlength{\leftmargin}{#1}}%
         \item[]%
 }
 {\end{list}}

\reportnumber{001} % Leave blank if n/a

\title{\centering RAGEN: Understanding Self-Evolution in LLM Agents via Multi-Turn Reinforcement Learning
}
\date{}

\author[*]{
Zihan Wang*$^1$, Kangrui Wang*$^1$, Qineng Wang*$^1$, Pingyue Zhang*$^1$, Linjie Li*$^{2}$, Zhengyuan~Yang$^4$, Xing Jin$^6$, Kefan Yu$^1$, Minh Nhat Nguyen$^7$, Licheng Liu$^1$, Eli Gottlieb$^1$, Yiping Lu$^1$, Kyunghyun~Cho$^5$, Jiajun Wu$^3$, Li Fei-Fei$^3$, Lijuan Wang$^4$, Yejin Choi$^3$, Manling Li$^{1}$
\\
\small $^1$Northwestern University~~~
$^2$University of Washington~~~
$^3$Stanford University~~~
$^4$Microsoft~~~\\
\small  $^5$New York University~~~
$^6$University of British Columbia~~~
$^7$Singapore Management University~~~\\
\vspace{-10pt}
  }
\correspondingauthor{$~$Equal Contribution.}

\renewcommand{\phi}{\varphi}

\renewcommand{\epsilon}{\varepsilon}
\renewcommand{\imath}{\mathrm{i}}

\newlength{\restsubwidth}
\newlength{\restsubheight}
\newlength{\restsubmoreheight}
\newcommand{\rest}[2]{%
        \settowidth{\restsubwidth}{\ensuremath{#2}}
        \settoheight{\restsubheight}{\ensuremath{{}_{#2}}}
        \ensuremath{{#1\hskip 0.5pt}_{\vrule\kern2pt\parbox[b][%
        4pt][b]{\the\restsubwidth}{%
                        \ensuremath{{}_{#2}}}}}
        }

\begin{abstract}
Training large language models (LLMs) as interactive agents presents unique challenges including long-horizon decision making and interacting with stochastic environment feedback. While reinforcement learning (RL) has enabled progress in static tasks, multi-turn agent RL training remains underexplored. We propose \textbf{StarPO} (\textbf{S}tate-\textbf{T}hinking-\textbf{A}ctions-\textbf{R}eward \textbf{P}olicy \textbf{O}ptimization), a general framework for trajectory-level agent RL, and introduce \textbf{RAGEN}, a modular system for training and evaluating LLM agents. 
Our study on four stylized environments reveals three core findings. First, our agent RL training shows a recurring mode of \textbf{Echo Trap} where reward variability cliffs and gradient spikes; we address this with \textbf{StarPO-S}, a stabilized variant with trajectory filtering, critic incorporation, and gradient stabilization.
Second, we find the shaping of RL rollouts would benefit from \textbf{diverse initial states, medium interaction granularity and more frequent sampling}. 
Third, we show that without \textbf{fine-grained, reasoning-aware reward signals}, agent reasoning hardly emerge through multi-turn RL and they may show shallow strategies or hallucinated thoughts.

\vspace{3pt}
\textbf{Keywords}: LLM Agents, Multi-turn RL \\
\textbf{Website}: \url{https://ragen-ai.github.io/} \\
\textbf{Code/Environments}:  \url{https://github.com/RAGEN-AI/RAGEN}. 

\end{abstract}

\begin{document}
\begin{CJK*}{UTF8}{gbsn}

\maketitle

\begin{figure}[ht]
    \centering
    \includegraphics[width=.93\linewidth]{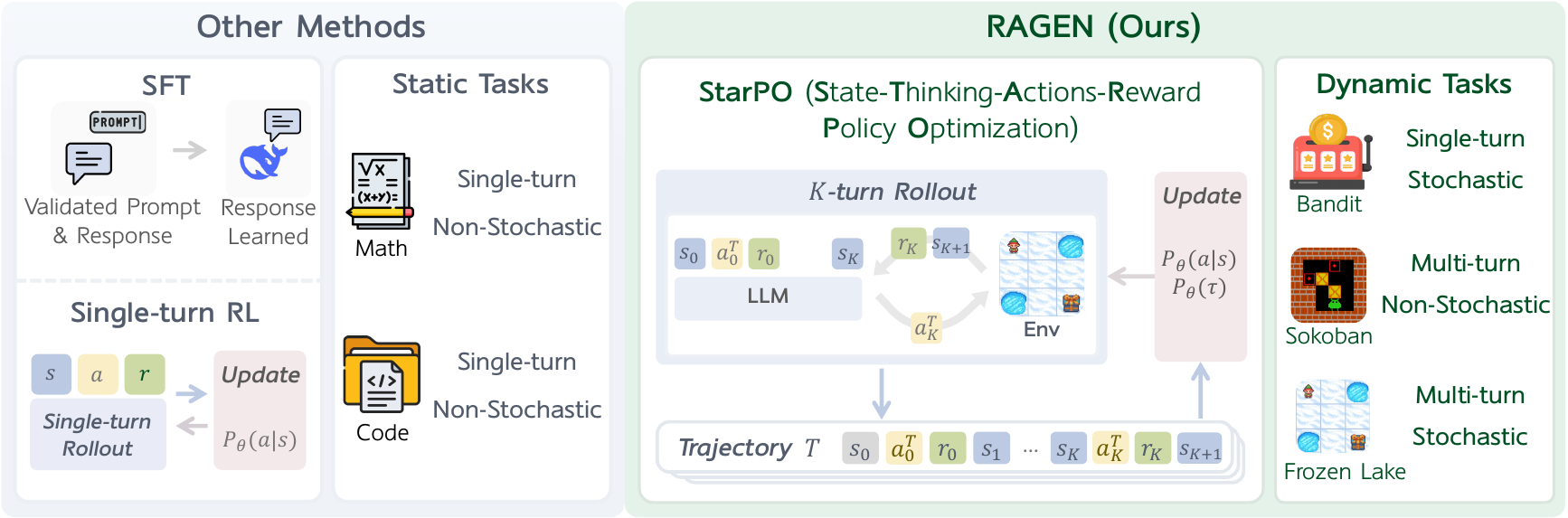}
    \caption{\small 
    % Comparison between RAGEN and existing LLM training frameworks. RAGEN provides a general reasoning agent training framework under an MDP formulation by leveraging the StarPO algorithm to optimize the entire trajectory distribution, learning multi-turn interactions in stochastic environments. 
    Previous methods focus on non-interactive tasks such as math or code generation. \textbf{RAGEN} implements StarPO, a general agent RL framework that supports multi-turn rollouts, trajectory-level reward assignment, and policy updates, on agent tasks requiring multi-turn stochastic interaction.  
    }
    \label{fig:teaser}
\end{figure}

\newpage

% \begin{spacing}{0.9}
% % \tableofcontents
% \end{spacing}

\newpage

\section{Introduction}

Training large language models (LLMs) to function as autonomous agents in interactive environments presents unique challenges. Unlike static tasks such as single-turn math problem solving~\citep{shao2024deepseekmathpushinglimitsmathematical} or coding~\citep{deepseekai2024deepseekcoderv2breakingbarrierclosedsource}, agent settings require models to make sequential decisions, maintain memory across turns, and adapt to stochastic feedback from their environment. These settings—central to planning assistants, robotics, and tutoring agents—demand that models not only perform well, but also self-improve through experience.

While recent work has explored reinforcement learning (RL) for LLMs~\citep{deepseekai2025deepseekr1incentivizingreasoningcapability, openai2024o1, tinyzero, zeng2025simplerl, kumar2024traininglanguagemodelsselfcorrect, gao2024regressingrelativefutureefficient} using rule-based reward, it remains largely underexplored to train interactive \textbf{agents} that self-evolve to reason and adapt through rule-based RL. In particular, LLM agent training often exhibits training instability, complex reward signals, and limited generalization across environment changes, especially under multi-turn interaction with stochastic feedback. A key open question is: \textit{what design factors make self-evolving LLM agents learn effectively and stably}?

We explore this question through a systematic study of agent learning under a general RL framework \textbf{StarPO} (\textbf{S}tate-\textbf{T}hinking-\textbf{A}ctions-\textbf{R}eward \textbf{P}olicy \textbf{O}ptimization). StarPO provides a unified view of \textbf{multi-turn, trajectory-level agent training} with flexible control over reasoning, reward assignment, and prompt-rollout structure. Built on top of StarPO, we develop \textbf{RAGEN}, a modular agent training and evaluation system designed to support the study of RL-based training in LLM Agents. RAGEN implements the full training loop—including rollout generation, reward assignment, and trajectory optimization—serving as a research infrastructure for systematic analysis of LLM agent training dynamics under multi-turn and stochastic environments.

% Training LLM agents on real-world tasks such as web browsing and embodied manipulation often relies on extensive pretrained priors and task-specific engineering.
% \textbf{To study learning from scratch and independent of these confounding factors}, we evaluate LLMs through RAGEN on four stylized environments: \textbf{Bandit} (single-turn, stochastic), \textbf{Sokoban} (multi-turn, deterministic), \textbf{Frozen Lake} (multi-turn, stochastic). These environments are deliberately minimalistic and fully controllable in difficulty, symbolic variation, and transition dynamics.
% Crucially, they require agents to learn decision-making policies through environment interaction, relying minimally on pre-existing world knowledge. The shared structure across these tasks (e.g., symbolic grid representations) further enables analysis of cross-task generalization.

Training LLM agents on real-world tasks like web browsing often depends on pretrained priors and heavy task-specific engineering. 
We evaluate RAGEN on four environments spanning different complexities: \textbf{Bandit} (single-turn, stochastic), \textbf{Sokoban} (multi-turn, deterministic), \textbf{Frozen Lake} (multi-turn, stochastic), and \textbf{WebShop} (multi-turn, open-domain).
The first three symbolic environments are \textbf{minimalistic and fully controllable}, while WebShop adds \textbf{real-world understanding and reasoning}. 
Together, they enable analysis of generalization across varied decision-making challenges.

Using this setup, we analyze three key dimensions of agent learning, and summarize below findings that \textbf{reveal core challenges and design principles} for stable agent RL training:
\vspace{-5pt}
\begin{enumerate}[left=2pt, itemsep=2pt, topsep=4pt]
    \item \textbf{Gradient Stability in Multi-turn RL is the Key to Stable Training.} We find that \textbf{multi-turn RL training} often leads to a recurring instability pattern, \textbf{Echo Trap}, where agents overfit to locally rewarded reasoning patterns, marked by reward variability collapse, entropy drop, and gradient spikes. To mitigate this failure mode, we propose \textbf{StarPO-S}, a stabilized variant of our framework that improves learning robustness through variability-based trajectory filtering, critic baselining, and decoupled clipping.
    
    \item \textbf{Rollout Frequency and Diversity Shape Self-Evolution.} In RL-based agent training, LLM self-generated rollout trajectories are served as core training material. We identify key rollout factors for stable agent RL training: (1) ensuring that rollouts come from \textbf{diverse initial states} with \textbf{multiple responses per initial state}, (2) \textbf{implementing multiple actions each turn} to improve interaction horizon within fixed turn limit, (3) maintaining a \textbf{high rollout frequency} to ensure online feedback reflects current policies. 
    % We find that training stability decreases when rollout lags behind updates and the model learns on old trajectories, or when prompt-response pairs lack intra-prompt comparison, highlighting the need for dynamic, structured rollout design.

    \item \textbf{Emerging \textit{Agent} Reasoning Requires Meticulous Reward Signal.} We find that simply encouraging reasoning in the action format does not guarantee reasoning behavior. Even when models are prompted to reason (e.g., with `\texttt{\textbf{<think>}}' tokens) with trajectory-level optimization via StarPO, they often regress to direct action selection if reasoning offers no distinct reward advantage. We assume this is due to the simple action spaces in MDP where shallow strategies suffice. Moreover, when rewards only reflect task success, models produce \textbf{hallucinated reasoning}, revealing a mismatch between thoughts and environment states. These issues underscore the need for \textbf{fine-grained, reasoning-aware reward design} for long-horizon agent training.

\end{enumerate}
\vspace{-5pt}

% Together, our framework and analysis offer insights into the principles behind training reasoning-capable, stable, and generalizable LLM agents. All environments and code are released as part of the RAGEN system.

\begin{figure*}[t]
    \centering
    \includegraphics[width=\textwidth]{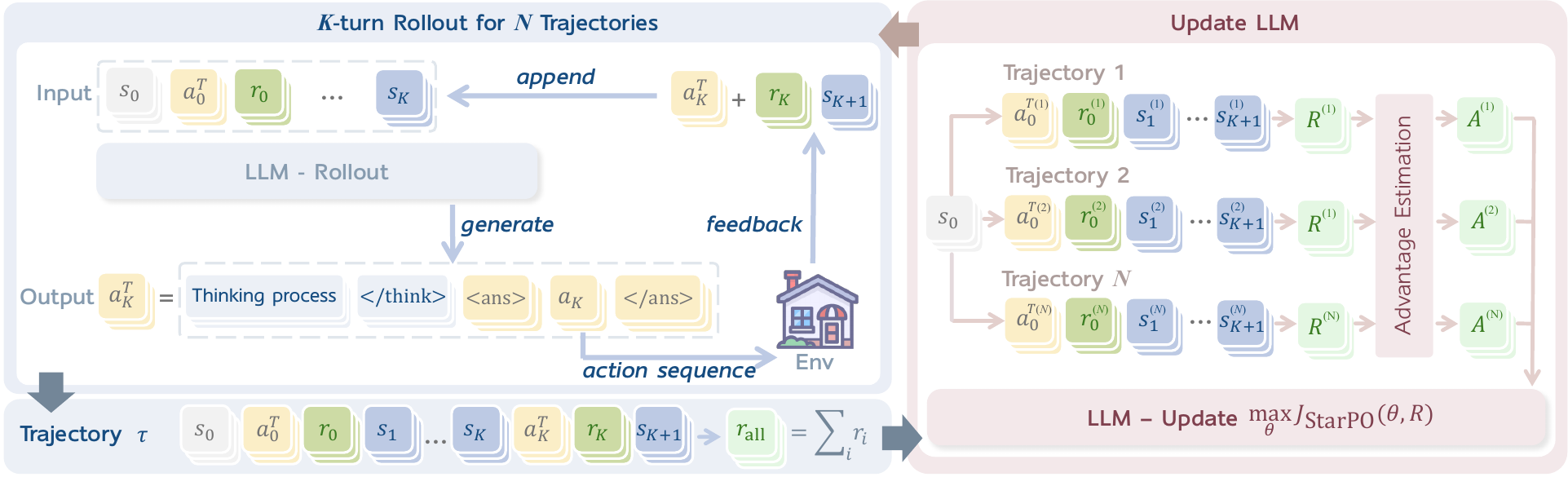}
    \caption{\small The State-Thinking-Actions-Reward Policy Optimization (StarPO) framework. LLM generates reasoning-guided actions for multi-turn interactions with environments and accumulates trajectory-level rewards, normalized and used to update the LLM policy.}
    \label{fig:starpo}
    \vspace{-5pt}
\end{figure*}

\section{Framework}

\subsection{The MDP Formulation for Agent Training}

Previous reinforcement learning (RL) for language models often assumes a single-turn setting, where the goal is to maximize the expected reward $R(s, a)$ over prompt-response pairs $(s, a)$ sampled from a dataset $\mathcal{D}$:
\begin{equation}
    J_{\text{step}}(\theta) = \mathbb{E}_{s \sim \mathcal{D}, a \sim \pi_\theta(\cdot|s)}[R(s, a)].
\end{equation}

However, LLM-based agents must operate in interactive environments that unfold over multiple turns and exhibit stochastic feedback. To capture these dynamics, we formulate the problem as a Markov Decision Process (MDP) $\mathcal{M} = \{S, A, P\}$, where
$S$ represents states (e.g., observation sequences or interaction histories), $A$ represents actions (often token sequences), and $P$ denotes the transition dynamics and reward generation process. The agent policy $\pi_\theta$ generates an action $a_t$ at each time step $t$, conditioned on the current state $s_t$ and the interaction history. The environment returns a reward $r_t$ and a new state $s_{t+1}$ given the current transition dynamics:
\[
    a_t \sim \pi_\theta(\cdot | s_t, \tau_{<t}), \quad (r_t, s_{t+1}) \sim P(\cdot | s_t, a_t),
\]
where $\tau_{<t} = \{s_0, a_0, r_0, ..., s_{t-1}, a_{t-1}, r_{t-1}\}$ denotes the interaction history. This interactive process continues for a maximum horizon $K$, yielding a full trajectory $\tau = \{s_0, a_0, r_0, ..., s_K\}$ that forms the learning material for the agent.

\subsection{StarPO: Reinforcing Reasoning via Trajectory-Level Optimization}

We introduce \textbf{StarPO} (State-Thinking-Action-Reward Policy Optimization), a general RL framework designed to optimize entire multi-turn interaction trajectories for LLM agents. Unlike previous methods for static tasks that treat each action independently, StarPO treats the \textbf{entire trajectory}—including observations, reasoning traces, actions, and feedback—as a coherent unit for rollout and model optimization.
The objective is to maximize expected trajectory reward:
\begin{equation}
    J_{\text{StarPO}}(\theta) = \mathbb{E}_{\mathcal{M}, \tau \sim \pi_{\theta}} \left[R(\tau) \right],
\end{equation}
where $\mathcal{M}$ is the MDP, $\tau$ is a full sequence of reasoning-augmented interactions, and $R(\tau)$ denotes the cumulative reward over the entire trajectory. The policy probability $\pi_\theta(\tau)$ is decomposed into token-level likelihoods, making StarPO directly compatible with autoregressive LLMs. Figure~\ref{fig:starpo} illustrates the full StarPO process, and we break them down in detail below.

\definecolor{bgcolor}{RGB}{232, 236, 240} 
\definecolor{badgecolor}{RGB}{181, 192, 208}  
\begin{onebox}{Trajectory-Level Objective in StarPO vs. Previous Methods}
\textbf{Previous methods (e.g., PPO~\citep{schulman2017proximalpolicyoptimizationalgorithms}, GRPO~\citep{shao2024deepseekmathpushinglimitsmathematical}):}
\[
\textcolor{black}{
J_{\text{step}}(\theta) = \mathbb{E}_{x \sim \mathcal{D},\, y \sim \pi_\theta(\cdot|x)}\left[ R(x, y) \right]
}
\quad
\textcolor{gray}{
\text{(optimize single-turn output } y \text{ given input } x)
}
\]

\textbf{StarPO (ours):}
\[
\textcolor{red!80!black}{
J_{\text{StarPO}}(\theta) = \mathbb{E}_{\mathcal{M}, \tau \sim \pi_\theta}\left[ R(\tau) \right]
}
\quad
\textcolor{red!80!black}{
\text{(optimize total reward over trajectory } \tau = \{s_0, a_0, r_0\dots, s_K\})
}
\]
\end{onebox}

\subsubsection{Optimization Procedure: Learning from Reasoning-Interaction Trajectories}

At each training iteration, the agent begins from an initial state $s_0$ and generates $N$ trajectories. At each step $t$, the agent produces a reasoning-guided structured output:
\begin{equation}
    a^T_t = \texttt{<think>}...\texttt{</think><answer>}~a_t~\texttt{</answer>},
\end{equation}
where $a^T_t$ is the full action output including intermediate reasoning, and $a_t$ is a sequence of environment-executable sub-action. The environment then returns the next state $s_{t+1}$ and reward $r_t$. The rollout stage produces complete trajectories $\tau = \{s_0, a^T_0, r_0, s_1, ..., a^T_{K-1}, r_{K-1}, s_K\}$, where \textit{every component is LLM-generated or environment-induced} and will be jointly optimized.

StarPO interleaves rollout and update steps.
% , supporting both online and offline learning modes.
New rollouts can be generated on-policy using $\pi_\theta$, or sampled from a replay buffer under $\pi_{\text{old}}$.
% This enables StarPO to flexibly unify online RL, offline RL, and hybrid training setups.
Each training loop consists of $P$ initial states $s_0$, each generating $N$ trajectories, and updates are performed with batch size $E$ for $L$ total loops. This yields $S = \frac{L \cdot P \cdot N}{E}$ total gradient update steps. Additional training mechanisms are discussed in \S\ref{sec:exp}.

\subsubsection{Modular Optimization Strategies}

StarPO supports a variety of policy optimization algorithms under a unified trajectory-level abstraction. For each rollout trajectory $\tau_i = \{\tau_{i,(1)}, \ldots, \tau_{i,(|\tau_i|)}\}$ of totally $|\tau_i|$ tokens, we instantiate StarPO with the following optimization strategies for token-level updates:
\vspace{-5pt}
\begin{itemize}[left=2pt, itemsep=2pt, topsep=4pt]
    \item \textbf{PPO~\citep{schulman2017proximalpolicyoptimizationalgorithms}.} We use the PPO objective (More details can be found in Appendix~\ref{app:a.rl}), where a critic is trained to estimate token-level value and advantages $A_{i,t}$:
    \begin{equation}
    \small
        J_{\text{PPO}}(\theta) = \frac{1}{G} \sum_{i=1}^G \frac{1}{|\tau_i|} \sum_{t=1}^{|\tau_i|}
        \min \left[
            \frac{\pi_\theta(\tau_{i,(t)}|\tau_{i,<t})}{\pi_{\text{old}}(\tau_{i,(t)}|\tau_{i,<t})} \cdot A_{i,t},\,
            \text{clip}\left( \frac{\pi_\theta(\tau_{i,(t)}|\tau_{i,<t})}{\pi_{\text{old}}(\tau_{i,(t)}|\tau_{i,<t})}, 1 - \epsilon, 1 + \epsilon \right) \cdot A_{i,t}
        \right],
    \end{equation}
    where $G$ is the number of trajectories in the batch, $\tau_{i,(t)}$ denotes the $t$-th token in trajectory $\tau_i$, and $\tau_{i,<t}$ is its prefix.

    \item \textbf{GRPO~\citep{shao2024deepseekmathpushinglimitsmathematical}.} For critic-free training leveraging GRPO, we assign a scalar reward $R(\tau_i)$ to each trajectory and normalized advantage $\hat{A}_{i,t}$ across all tokens in $\tau_i$:
    \begin{equation}
    \small
        \hat{A}_{i,t} = \frac{R(\tau_i) - \text{mean}(\{R(\tau_1), \ldots, R(\tau_G)\})}
                             {\text{std}(\{R(\tau_1), \ldots, R(\tau_G)\})}.
    \end{equation}
    The GRPO objective becomes:
    \begin{equation}
        \small
        J_{\text{GRPO}}(\theta) = \frac{1}{G} \sum_{i=1}^G \frac{1}{|\tau_i|} \sum_{t=1}^{|\tau_i|} 
        \min \left[
            \frac{\pi_\theta(\tau_{i,(t)}|\tau_{i,<t})}{\pi_{\text{old}}(\tau_{i,(t)}|\tau_{i,<t})} \cdot \hat{A}_{i,t},\,
            \text{clip}\left( \frac{\pi_\theta(\tau_{i,(t)}|\tau_{i,<t})}{\pi_{\text{old}}(\tau_{i,(t)}|\tau_{i,<t})}, 1 - \epsilon, 1 + \epsilon \right) \cdot \hat{A}_{i,t}
        \right].
    \end{equation}
\end{itemize}

\subsection{The RAGEN System}

To implement StarPO in practice, we build \textbf{RAGEN}, a complete system for LLM agent training in controlled environments. RAGEN supports structured rollouts, customizable reward functions, and integration with multi-turn, stochastic environments. It serves both as the execution backend for StarPO and as a platform for studying stability, generalization, and learning dynamics in training reasoning agents.
RAGEN is designed to be extensible: new environments, reward schemes, or rollout  strategies can be easily plugged in, serving as a foundation for RL-based agent training.

\vspace{-2pt}
\section{Experiment Setup}
\label{sec:exp}
\vspace{-2pt}
\subsection{Environments and Tasks}
\label{sec:envs}
% We evaluate LLM agents in three symbolic environments designed to study learning from scratch and independent of task-specific engineering. All environments are minimal, fully controllable, and stripped of real-world priors to support clean analysis. Specifically, \textbf{Bandit} tests risk-sensitive symbolic reasoning under stochastic feedback; \textbf{Sokoban} requires irreversible multi-step planning in a deterministic setting; and \textbf{Frozen Lake} combines planning with probabilistic transitions. Visualizations and environment illustrations are deferred to Appendix~\ref{app:env-setting}. We also include experiments on more realistic task, WebShop~\cite{yao2022webshop} and include the results in Appendix~\ref{app:webshop}.

We evaluate LLM agents on four environments spanning symbolic and realistic decision-making: \textbf{Bandit} tests risk-sensitive reasoning under noisy feedback; \textbf{Sokoban} requires irreversible symbolic planning; \textbf{Frozen Lake} combines planning with probabilistic transitions; and \textbf{WebShop} involves natural language grounding and web environment interaction. 
The first three symbolic environments are deliberately minimal and fully controllable to support clean analysis, while WebShop introduces realistic task structure and language input. Environment visualizations are in Appendix~\ref{app:env-setting}.
\vspace{-5pt}

\subsection{Training Settings}
In our main experiments, we train Qwen-2.5 Instruct 0.5B models for three symbolic tasks and its 3B variant for the challenging WebShop. We also report various model performance in Appendix~\ref{app:largermodelabaltion}. Models are trained with StarPO variants on H100 GPUs for 100-200 rollout–update iterations. Each batch samples $P{=}8$ prompts, with $N{=}16$ rollouts per prompt, up to 5 turns and 10 actions. Policy updates use GRPO or PPO with GAE ($\gamma{=}1.0, \lambda{=}1.0$), Adam optimizer, entropy bonus ($\beta{=}0.001$), and a response-format penalty ($-0.1$). More details can be found in Appendix~\ref{app:train-eval-setting}.
\vspace{-5pt}

\subsection{Evaluation Metrics}
We evaluate on 256 fixed prompts per environment with temperature $T{=}0.5$, truncating episodes after 5 turns. Metrics include: 
\textbf{(i)} success rate (task completion),
\textbf{(ii)} rollout entropy (exploration),
\textbf{(iii)} in-group reward variability (behavioral diversity),
\textbf{(iv)} response length (reasoning verbosity),
and \textbf{(v)} gradient norm (training stability).
All are computed over validation instances. More details can be found in Appendix~\ref{app:metric-setting}.
\vspace{-5pt}

\section{Experimental Results and Findings}
\subsection{Multi-turn Agent RL Training Introduces New Instability Pattern}
\vspace{-5pt}

\begin{figure}[t]
  \centering
  \includegraphics[width=\linewidth]{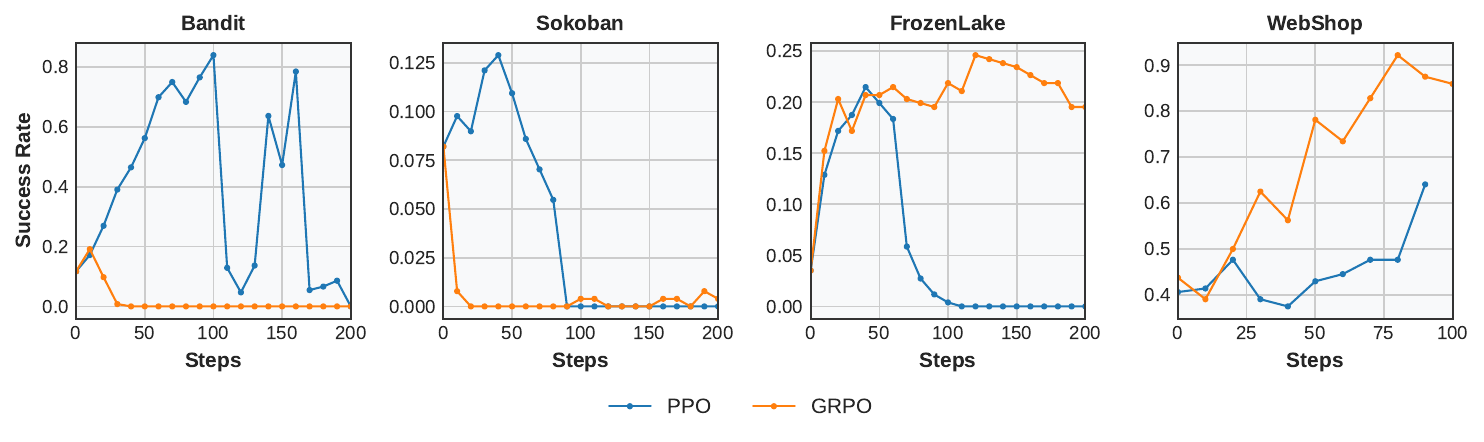}
    \caption{
    \textbf{Baseline StarPO performance across environments.}
     Symbolic tasks like Bandit and Sokoban lead to collapse while real-world WebShop starts high and improves fast. 
     PPO is stronger in Bandit and Sokoban to provide more stable token-level reward signal, while GRPO is stronger in Frozen Lake whose randomness makes the state value hard to estimate, and WebShop whose strong initial performance lessens the need for a critic to stabilize gradients.
    }
  \label{fig:task_performance}
  \vspace{-5pt}
\end{figure}

We evaluate baseline StarPO across agent tasks (Figure~\ref{fig:task_performance}). Symbolic environments like Bandit and Sokoban show early improvements but eventually collapse. PPO is more stable than GRPO in these settings, collapsing later and achieving higher performance, likely due to its critic providing smoother reward estimates. Interestingly, GRPO is more stable on Frozen Lake, likely due to the difficulty of estimating state values in this task, which may destabilize PPO (see Appendix~\ref{app:frozenlake-grpo}). On WebShop, both methods succeed, likely due to strong language prior and high initial rewards reducing the need for a critic.

% \begin{figure}[t]
%   \centering
%   \includegraphics[width=0.32\linewidth]{figures/0_bandit_performance.pdf}
%   \includegraphics[width=0.32\linewidth]{figures/0_sokoban_performance.pdf}
%   \includegraphics[width=0.32\linewidth]{figures/0_frozenlake_performance.pdf}
%     \caption{
%     \textbf{Baseline StarPO performance across environments.}
%     All StarPO runs initially improve but eventually collapse in multi-turn agent settings. PPO variant shows better training stability compared to the GRPO variant, especially in Bandit and Sokoban, indicating that critic-based methods better resist early-stage degradation under long-horizon dynamics.
%     }
%   \label{fig:task_performance}
% \end{figure}

\newcolumntype{L}[1]{>{\raggedright\arraybackslash\small\ttfamily}p{#1}}

\begin{onebox}{Finding 1: Single-turn RL may not be directly adapted to Multi-turn agent RL}
Vanilla adaptations from single-turn methods like PPO and GRPO achieve early gains in agent settings but often collapse. A critic in PPO may delay instability, but would not prevent reasoning degradation, highlighting the need for specialized stabilization in agent settings.
\end{onebox}

To understand the causes of collapse, we compare early- and late-stage trajectories. In the Bandit task, early-stage trajectories exhibit diverse reasoning about symbolic meanings and expected rewards, while later-stage responses become repetitive and deterministic. This suggests that \textbf{RL training may have over-amplified inherent reasoning shortcuts}, reinforcing locally rewarded templates while suppressing exploration. 
We refer to this failure mode as an “\textbf{Echo Trap}”, which is similar to the findings in~\citet{recursivelygenerated}, where the model repeatedly reuses memorized reasoning paths when trained on self-generated trajectories, leading to a collapse in diversity and long-term performance degradation. Examples are in Appendix~\ref{app:echotrap}.

\begin{onebox}{Finding 2: Model collapse in agent RL is reflected as "Echo Trap" over training}
We find that early-stage agent respond with diverse symbolic reasoning, but collapse into deterministic, repetitive templates after training. Models converge to fixed phrasing, indicating that RL may reinforce superficial patterns instead of general reasoning and forms an "Echo Trap" that hinders long-term generalization.
\end{onebox}

To detect collapse, we monitor two key metrics: (1) \textbf{Average Reward}, where a plateau or drop signals degraded task performance, and (2) \textbf{Gradient Norm}, where spikes suggest unstable updates. To identify early signs, we also track rollout-level signals: (1) \textbf{Reward Standard Deviation}, indicating whether the policy is producing diverse outcomes or collapsing into repetitive behavior, and (2) \textbf{Output Entropy}, which indicates how deterministic the model’s predictions are.

% To determine whether collapse has already occurred, we track two metrics: (1) \textbf{Average Reward} whose plateau or drop identifies degraded task-solving ability, (2) \textbf{Gradient Norm} that measures update magnitude, with its spikes indicate that small updates are triggering large changes, typically a sign of training instability and collapse.

% To identify early signs of collapse, we further analyze rollout-level statistical signals that reflect behavioral redundancy and policy instability. Specifically, we examine two early indicators:
% (1) \textbf{Reward Standard Deviation} that measures whether rollouts under the same task instance lead to diverse outcomes or has collapsed into repetitive patterns with similar returns, (2) \textbf{Output Entropy} that captures the uncertainty of the model’s output, whose sharp drop implies the policy is becoming overconfident and converging to narrow reasoning paths. 

\begin{figure}[t]
  \centering
  \includegraphics[width=0.7\textwidth]{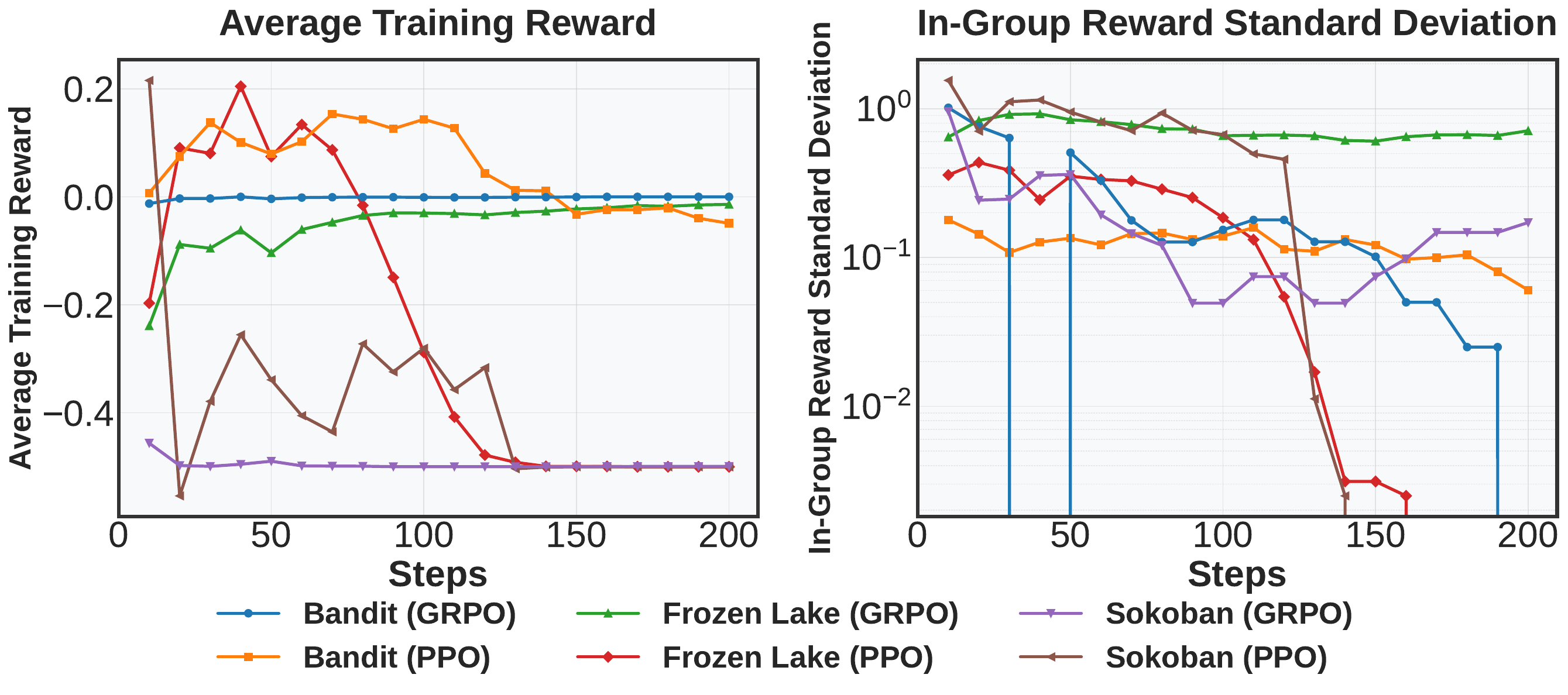}
  \includegraphics[width=\textwidth]{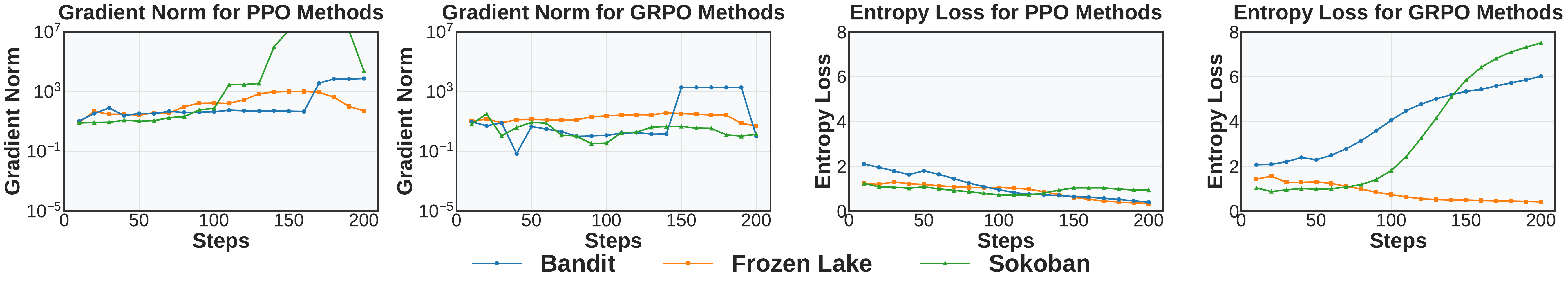}
  \caption{
\textbf{Collapse indicators and early warning signals in multi-turn RL.}
Average reward and gradient norm (left-side plots) reflect collapse directly, and their plateaus and spikes confirm performance and training instability.
Reward standard deviation and entropy (right-side plots) often becomes unstable before reward degrades, serving as early warning signals. 
}
  \label{fig:collapse_metrics}
  \vspace{-5pt}
\end{figure}

% Figure~\ref{fig:collapse_metrics} summarizes these dynamics across tasks and optimization methods. From the results, we summarize key conclusions regarding \textbf{how model collapse happens}:
% \vspace{-5pt}
% \begin{itemize}
%     \item \textbf{Reward Std is a strong predictor for model collapse in agent RL.} For FrozenLake-PPO, reward mean collapses at 90 steps, but std drops sharply as early as 40 when the reward mean reaches the peak—serving as an early warning. In Bandit-PPO, std reaches its lowest point around 70 steps, just as reward peaks at 120. In Sokoban-PPO, std and mean collapse almost simultaneously around step 10.
%     \item \textbf{Gradient norm spikes mark the point of complete collapse.} After gradient norm spikes in PPO appear at 170 (Bandit), 110 (Sokoban), and 90 (FrozenLake), it basically means even small updates cause destabilizing shifts, rendering the model unable to recover after acute update steps.
%     
% \end{itemize}
% \vspace{-5pt}

Figure~\ref{fig:collapse_metrics} summarizes these dynamics across tasks and optimization methods. From the results, we draw the following conclusions regarding \textbf{how model collapse emerges in multi-turn agent RL}:
\vspace{-5pt}
\begin{itemize}[left=2pt, itemsep=2pt, topsep=4pt]
    \item \textbf{Reward standard deviation is an early indicator of convergence.} In FrozenLake-PPO, std drops sharply at step 40, well before reward mean collapses at step 90 while performance is near-optimal. In Bandit-PPO, std bottoms out around step 70, preceding the reward peak at step 120. In Sokoban-PPO, std and mean collapse together near step 10, suggesting early saturation.

    \item \textbf{Gradient norm spikes indicate irreversible collapse.} Once gradient norm spikes emerge, e.g., at step 170 (Bandit), 110 (Sokoban), and 90 (FrozenLake), even small parameter updates could induce drastic loss shifts, after which recovery becomes unlikely.
    
    \item  \textbf{Entropy should follow a stable decay trend during effective learning}. This can be seen from FrozenLake-GRPO. Rapid entropy increases or erratic changes often correlate with collapsed reasoning behavior, such as GRPO on Bandit and Sokoban.
\end{itemize}
\vspace{-9pt}

\begin{onebox}{Finding 3: Collapse follows similar dynamics and can be anticipated by indicators}
\textbf{Reward standard deviation} and \textbf{entropy} often fluctuate before performance degrades, while \textbf{gradient norm} spikes typically mark the point of irreversible collapse. These metrics provide early indicators and motivate the need for stabilization strategies.
\end{onebox}
These patterns confirm that multi-turn RL introduces unique challenges that single-turn RL methods fail to handle. In response, we introduce \textbf{StarPO-S}, a stabilized variant that targets sampling quality, gradient stability, and exploration regularization to avoid premature collapse.

\subsection{StarPO-S: Stabilize Multi-turn RL with Instance Filtering and Gradient Shaping}

To address the instability of multi-turn reinforcement learning, we introduce \textbf{StarPO-S}, a stabilized variant of StarPO that incorporates three key modifications aimed at improving training robustness and efficiency. Building on the insight that declining reward standard deviation often precedes collapse, we investigate the following question: \textit{should agents be trained more intensively on task instances where their behavior is more uncertain with higher reward variability?} 

\begin{figure}[t]
  \centering
  \includegraphics[width=\textwidth]{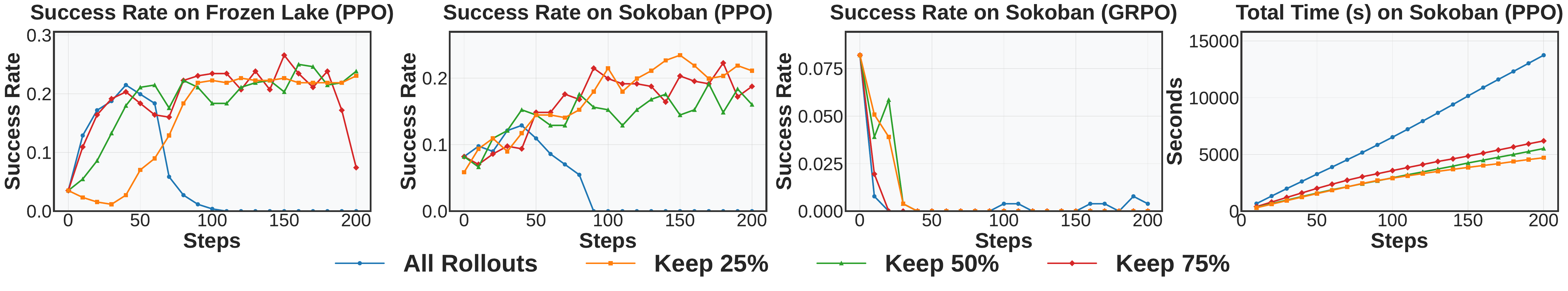}
  \caption{\textbf{Effect of uncertainty-based filtering on multi-turn RL stability.} Filtering out low-variability trajectories reduces collapse risk and improves success rate. On PPO variants, collapse is largely mitigated when more than half of the trajectories are filtered. Training time is reduced.}
  \label{fig:filtering_starpos}
  \vspace{-5pt}
\end{figure}

We hypothesize that the most effective training samples are those where the agent \textbf{exhibits outcome uncertainty}—avoiding both trivial task instances and overly difficult ones. This intuition is rooted in principles of Active Learning~\citep{activelearning}, where uncertain examples are the most informative ones models should learn from. We define trajectory-level outcome uncertainty $U$ for policy $\pi_\theta$ on a given agent task instance (initial state $s_0$ in an MDP $\mathcal{M} = \{S, A, P\}$) as:
\begin{equation}
\text{U}(\pi_{\theta}, \mathcal{M}, s_0) = \text{Std}_{\tau \sim \pi_\theta(\cdot | s_0)}\left[R(\tau)\right].
\end{equation}
During training, we sort prompts based on the standard deviation of reward obtained from repeated rollouts and \textbf{retain only the top $p$\% highly-uncertain prompts} at each training step. Figure~\ref{fig:filtering_starpos} shows the effect of varying $p$ in PPO and GRPO under StarPO-S. We further validate the effect of uncertainty-based filtering in Appendix~\ref{app:std}.

In PPO runs (Figure~\ref{fig:filtering_starpos}, left), filtering low-variability rollouts significantly delays collapse: retaining 75\% of rollouts extends stability in FrozenLake from 100 to 140 steps, while 50\% avoids collapse entirely. GRPO remains less stable due to its critic-free design but still benefits modestly. Filtering also improves efficiency (Figure~\ref{fig:filtering_starpos}, right). We adopt 25\% as the default for StarPO-S. However, we note that this aggressive value may not be optimal for all scenarios.
Tasks like Sokoban and FrozenLake respond well to aggressive filtering, potentially due to their relatively repetitive reasoning patterns and under-representation in pretraining, which make them tend to collapse when similar trajectories dominate the batch. We further present larger model (72B), together with frontier model such as GPT-4o and Qwen-2.5-72B, in Appendix~\ref{app:largermodelabaltion}, to better contextualize our model performance.

\begin{onebox}{Finding 4: Filtering low-variance trajectories improves stability and efficiency}
Training on high-variance prompts delays or eliminates collapse in multi-turn RL. StarPO-S improves performance and reduces update steps by discarding low-information rollouts, especially under PPO. This aligns with active learning principles, where uncertain examples offer the most informative learning signals.
\end{onebox}

In addition to uncertainty-based filtering, we adopt two gradient shaping techniques inspired by DAPO~\citep{yu2025DAPO} designed for single-turn RL: \textbf{KL Term Removal} and \textbf{Clip-Higher} (Asymmetric Clipping). We extend and evaluate them in the multi-turn agent setting, and find both methods boost the success rate and extend stable training phases, showing how multi-turn RL benefits from more flexible gradient shaping. Design details and performance ablation is in Appendix~\ref{app:largermodelabaltion}.

\begin{figure}[t]
  \centering
  \includegraphics[width=\textwidth]{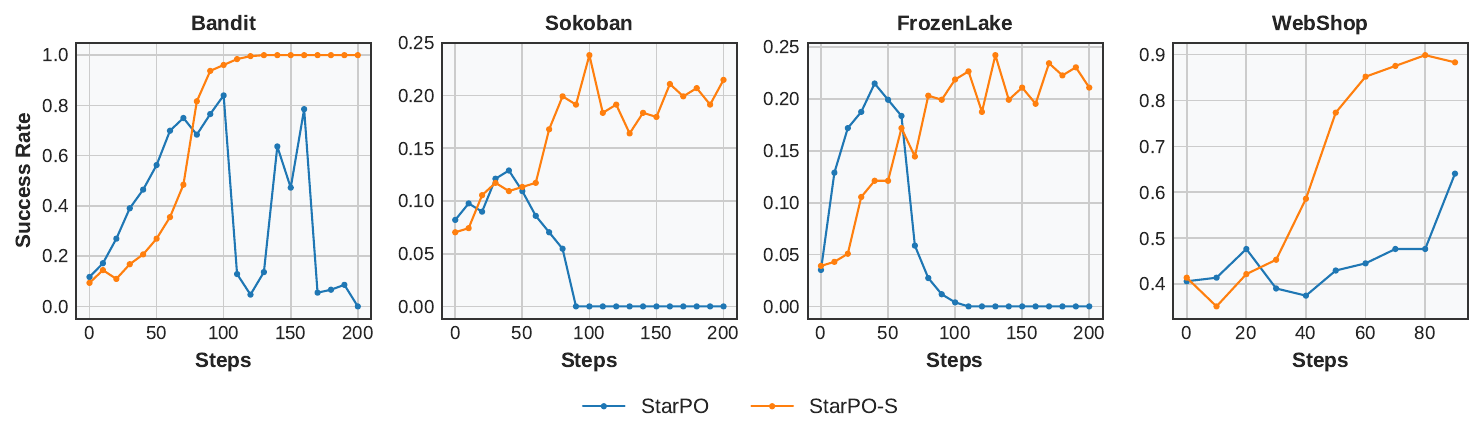}
  \caption{\textbf{StarPO-S improves stability and final performance across tasks.} Compared to vanilla StarPO, StarPO-S reliefs collapse in all four tasks and could achieves higher success rates. }
  \label{fig:starpos_summary}
  \vspace{-9pt}
\end{figure}

\noindent\textbf{Overall Comparison.}~~
    We compare StarPO-S with vanilla StarPO across three tasks in Figure~\ref{fig:starpos_summary}. StarPO-S consistently delays collapse and enhances final task performance. We attribute these gains to more selective training data (via uncertainty filtering), more balanced optimization signals (via KL removal and decoupled clipping), reducing narrowed reasoning modes. In the appendix~\ref{app:largermodelabaltion}, we further discuss about other variants that may stablize training and enhance performance, such as selective response mask and Bi-level Gegenal Advantage Estimation (GAE)~\cite{vagen}.

% In addition to filtering, we incorporate two enhancements adapted from DAPO~\citep{yu2025DAPO} for stabilizing single-turn RL, and evaluate their effect in multi-turn settings:

% \begin{itemize}
%     \item \textbf{KL Removal:} We remove the KL divergence penalty from PPO's loss function, relying solely on policy loss and entropy bonus to shape the update.
%     \item \textbf{Clip-High (Decoupled Clipping):} We apply asymmetric clipping ranges in PPO: a higher upper bound (e.g., $\epsilon_\text{high} = 0.4$) than the lower bound ($\epsilon_\text{low} = 0.1$), allowing larger gradient updates on high-reward trajectories while suppressing punishment for wrong but low-reward ones.
% \end{itemize}

% \begin{figure}[H]
%   \centering
%   \includegraphics[width=0.48\textwidth]{figures/3_sokoban_comparetrick_ppo.pdf}
%   \includegraphics[width=0.48\textwidth]{figures/3_frozenlake_comparetrick_ppo.pdf}
%   \caption{Effect of KL removal and asymmetric clipping (ClipHigh) in PPO. Both improve peak performance and delay collapse.}
%   \label{fig:trick_results}
% \end{figure}

% As seen in Figure~\ref{fig:trick_results}, both methods increase the peak success rate and extend training stability in PPO. This supports the idea that multi-turn RL agents benefit from more selective gradient shaping—focusing on amplifying effective reasoning paths rather than aggressively penalizing noisy ones.

\subsection{Generating Useful Trajectories for RL Training}
\label{sec:rollout}
Effective RL training depends heavily on trajectory quality. We study three key rollout dimensions—\textit{task diversity}, \textit{interaction granularity}, and \textit{rollout frequency}—by training on vanilla Sokoban and evaluating on other tasks including SokobanNewVocab, LargeSokoban, and FrozenLake Task, which we detail in Appendix~\ref{app:generalization-task}.

\noindent\textbf{Higher task diversity with response comparison improves generalization.}~~
Task diversity refers to the number of distinct prompts used for each rollout-update cycle. With a fixed batch size, it trades off against the number of responses per prompt. 
In our experiments (Table~\ref{tab:diversity}), we vary this trade-off and find that higher task diversity—achieved by fewer responses per prompt (e.g., 4 per prompt)—consistently yields better generalization. This only holds when each prompt includes multiple rollouts, enabling the agent to contrast different outcomes under similar conditions.

\noindent\textbf{Allowing more action budgets enables planning, while overly long-horizon rollouts inject noise.}~~
We vary the number of actions allowed per turn in Table~\ref{tab:granularity}. Allowing up to 5 or 6 actions per turn yields the best performance, especially on complex environments like SokobanNewVocab and LargeSokoban. This setting provides enough room for planning while avoiding the chaos of overly long rollouts. Increasing the budget to 7 actions degrades performance, likely due to noisy transitions and diluted reward feedback.

\begin{table}[t]
  \centering
  \small
    \caption{\textbf{Effect of Task Diversity on Generalization Performance (\%).} Higher diversity with multiple responses yields the best performance (4 responses per prompt).}
  \begin{tabular}{lccc}
    \toprule
    \textbf{Response Per Prompt} & \textbf{SingleSokoban} & \textbf{SokobanNewVocab} & \textbf{FrozenLake} \\
    \midrule
    32   & 21.09\% & 20.22\% & 17.97\% \\
    16   & 20.31\% & 21.48\% & 19.53\% \\
    8  & 20.31\% & 19.53\% & 17.19\% \\
    4  & \textbf{20.70\%} & \textbf{25.39\%} & \textbf{21.48\%} \\
    2  & 19.92\% & 25.00\% & \textcolor{red}{12.50\%} \\
    1 & 19.53\% & 22.27\% & \textcolor{red}{12.50\%} \\
    \bottomrule
  \end{tabular}
    \vspace{-8pt}
  \label{tab:diversity}
\end{table}

\begin{table}[t]
  \centering
  \small
  \caption{\textbf{Performance across environments under different per-turn action budgets (\%).} 5--6 actions per turn yields best performance, balancing well for effective multi-step planning.}
  \begin{tabular}{lcccc}
    \toprule
    \textbf{Max Actions / Turn} & \textbf{Sokoban} & \textbf{SokobanNewVocab} & \textbf{LargeSokoban} & \textbf{FrozenLake} \\
    \midrule
    1 & 12.11\% & 13.67\% & 1.17\% & 11.72\% \\
    2 & 16.41\% & 21.09\% & 3.52\% & 18.36\% \\
    3 & 19.53\% & 19.53\% & 1.95\% & 20.88\% \\
    4 & 26.95\% & 26.95\% & 5.08\% & 20.70\% \\
    5 & 28.13\% & 25.78\% & 6.25\% & \textbf{21.09\%} \\
    6 & \textbf{33.59\%} & \textbf{31.64\%} & \textbf{6.64\%} & \textcolor{red}{18.36\%} \\
    7 & \textcolor{red}{22.27\%} & 28.52\% & \textcolor{red}{3.91\%} & \textcolor{red}{19.53\%} \\
    \bottomrule
  \end{tabular}
  \label{tab:granularity}
\vspace{-10pt}
\end{table}

\begin{table}[t]
  \centering
  \small
  \caption{\textbf{Generalization performance (\%) with and without reasoning under StarPO-S.} Disabling reasoning significantly reduces generalization in single-turn Bandit task, but has mixed or marginal effects in multi-turn Sokoban task.}
  \begin{tabular}{lcc|cccc}
    \toprule
    & \multicolumn{2}{c|}{\textbf{Train on Bandit}} & \multicolumn{4}{c}{\textbf{Train on Sokoban}} \\
    \cmidrule(r){2-3} \cmidrule(l){4-7}
    & Bandit & Bandit-Rev & FrozenLake & LargeSokoban & Sokoban & SokobanNewVocab \\
    \midrule
    StarPO-S & \textbf{100.00} & \textbf{67.58} & \textbf{19.92} & 2.34 & 21.48 & 18.75 \\
    NoThink & 81.25 & 56.25 & 19.53 & \textbf{2.73} & \textbf{20.73} & \textbf{26.17} \\
    \bottomrule
  \end{tabular}
\label{tab:reasoning_generalization_grouped}
\vspace{-8pt}
\end{table}

\begin{figure}[H]
  \centering
  \includegraphics[width=\textwidth]{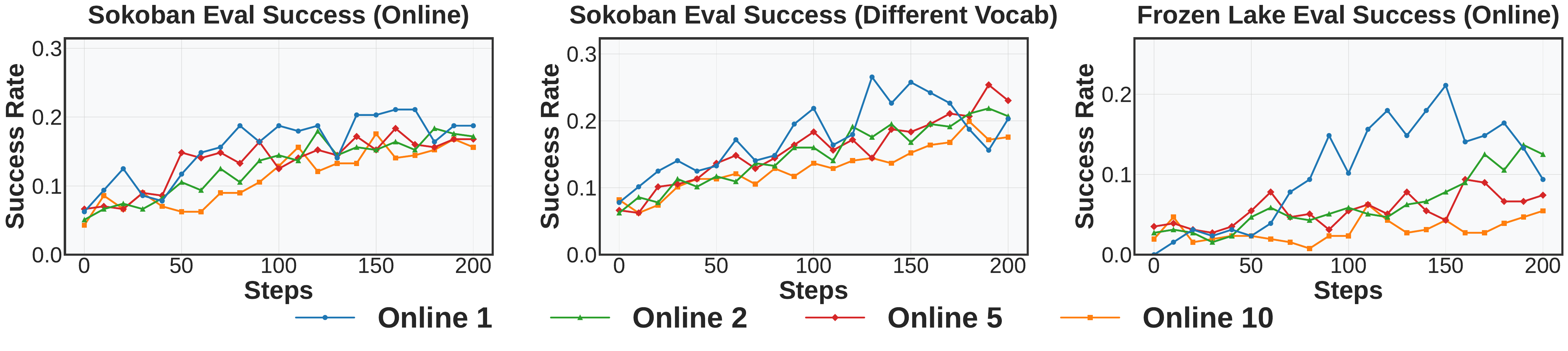}
\caption{\textbf{Performance under different rollout frequencies (\textit{Online-\(k\)}).} We vary the rollout reuse factor \(k\), where each batch is reused for \(k\) policy updates. Lower \(k\) (e.g., \textit{Online-1}) means more frequent rollouts. Fresher data improves convergence by staying aligned with the current policy.} 
  \label{fig:rollout_freq}
\vspace{-15pt}
\end{figure}

\noindent\textbf{Frequent rollout updates ensure alignment between optimization targets and current policy behavior.}~~
To investigate the effect of rollout freshness, we adopt an \textit{Online-\(k\)} rollout strategy, where a single set of rollouts is reused for \(k\) consecutive policy updates. A smaller \(k\) implies more frequent rollout collection. Notably, \textit{Online-1} corresponds to an fully online setting, with fresh rollouts collected every update iteration. 
As shown in Figure~\ref{fig:rollout_freq}, agents trained with fresher rollouts (\textit{Online-1}) achieve faster convergence and better generalization across tasks compared to those with delayed updates (e.g., \textit{Online-5} or \textit{Online-10}). This supports a core design principle for multi-turn RL: learning is most effective when trajectories reflect the agent’s latest behavior. Frequent rollout reduces policy-data mismatch and improves optimization stability.

\begin{onebox}{Finding 5: Task diversity,  action budget, and rollout frequency affect rollout quality}
Diverse task instances enable better policy contrast and generalization across environments.
Moderate action budgets provide enough planning space and avoid the noise introduced by overly long sequences. 
Up-to-date rollouts ensure optimization targets remain aligned with current policy behavior. 
\end{onebox}

\subsection{Reasoning Improves Generalization but Fades in Multi-Turn Settings Without Fine-Grained Rewards}

We examine how symbolic reasoning impacts agent generalization. While reasoning enhances performance in single-turn tasks like Bandit, it fails to grow or sustain in complex multi-turn environments like Sokoban. Below, we analyze these effects step-by-step.

\noindent\textbf{Reasoning traces improve generalization in single-turn Bandit tasks.}~~
We design a controlled generalization test in symbolic Bandit environments. In the original \texttt{Bandit} setting, the model is trained on the \texttt{[Teacher, Engineer]} arm pair and evaluated on \texttt{[Librarian, Trader]}, preserving intuitive risk-reward alignments (i.e., \texttt{Engineer} and \texttt{Trader} are high-risk, high-reward). In \texttt{BanditRev}, these associations are inverted, assigning counter-intuitive reward profiles to professions and making reasoning more challenging.

As shown in Table~\ref{tab:reasoning_generalization_grouped}, models trained with reasoning traces generalize better in \texttt{Bandit} and even in the counterintuitive \texttt{BanditRev}, suggesting that reasoning supervision helps internalize symbolic cues beyond memorization.
Despite the added difficulty in \texttt{BanditRev}, models with explicit reasoning consistently outperform those without, as shown in Table~\ref{tab:reasoning_generalization_grouped}. This suggests that reasoning traces help the agent internalize symbolic-reward associations and generalize beyond surface-level memorization, even under semantic-reward misalignment.

\noindent \textbf{In multi-turn tasks, reasoning signals fade as training progresses.}~~ In contrast to single-turn settings, we find that reasoning provides limited benefits in multi-turn environments like Sokoban and FrozenLake. Even when the output format includes explicit \texttt{<think>} segments, removing them (no-think variant) often yields comparable or even better performance. To understand this degradation, we analyze average response length during training (Table~\ref{tab:reason_length}, Figure~\ref{fig:reasoning_decay}) and find that reasoning traces consistently shrink over time, suggesting the model is suppressing its own thought process. Interestingly, in the semantically misaligned \texttt{BanditRev} task—where reasoning is essential—traces remain longer, indicating that reasoning is better sustained when context is more challenging.

We presume that reasoning collapse may arise from the \textbf{sparse, delayed reward structure in multi-turn tasks}, which often fails to differentiate between coherent reasoning and trial-and-error success. This is supported by examples (Appendix~\ref{app:case-study-spurious}) where models generate incoherent or hallucinated reasoning yet still receive high rewards. This raises an important challenge: \textit{how can we consistently reinforce useful reasoning when the reward alone may not reflect its quality?} One possible approach is to decouple action correctness from reasoning quality using format-based penalties: similar to~\citet{deepseekmath}, we apply a small penalty to outputs lacking valid \texttt{<think>}–\texttt{<answer>} structure, encouraging structured reasoning. We believe future work may explore finer-grained reward designs such as rewarding partial correctness to reliably reinforce reasoning in long-horizon decision-making.

\begin{table}[t]
\centering
\small
\caption{\textbf{Reasoning length (\texttt{<think>} block length) at different training steps.} Token length generally declines over time, while contextually challenging problems like \texttt{ReverseBandit} require more reasoning than their original counterparts.}
\vspace{2pt}
\begin{tabular}{lcccc|cc}
\toprule
 & \multicolumn{4}{c|}{\textbf{Bandit}} & \multicolumn{2}{c}{\textbf{Sokoban}} \\
\textbf{Step} & Original & NoThink & Reverse & Reverse NoThink & Original & NoThink \\
\midrule
0   & 66.0 & 12.7 & 68.8 & 12.7 & 307.1 & 68.6 \\
100 & 25.6 & 12.4 & 33.7 & 13.0 & 104.6 & 55.3 \\
200 & 17.6 & 12.4 & 30.7 & 13.0 & 89.5  & 60.0 \\
\bottomrule
\end{tabular}
\label{tab:reason_length}
\vspace{-15pt}
\end{table}

\begin{onebox}{Finding 6: Reasoning fails to emerge without meticulous reward design}
While symbolic reasoning can emerge in simple, single-turn tasks under weak supervision, it fails to persist in multi-turn environments without the reward design explicitly encouraging interpretable intermediate reasoning steps. We observe that even with structured prompts, reasoning gradually decays during training if the reward signal focuses only on final outcomes. This suggests that without meticulous reward shaping, agents may tend to collapse into shortcut behaviors that bypass reasoning altogether.
\end{onebox}

\section{Related Work}
Recent work has explored fine-tuning LLMs for multi-step reasoning through \textbf{reinforcement learning (RL)}, and developing agent frameworks to structure decision-making tasks. For reasoning, techniques range from classical PPO~\citep{schulman2017proximalpolicyoptimizationalgorithms} and actor-critic methods~\citep{haarnoja2018softactorcritic} to structured prompting via meta tokens~\citep{goyal2024thinkspeaktraininglanguage, herel2024thinkingtokenslanguagemodeling}. Policy variants such as RLOO~\citep{rloo}, GRPO~\citep{deepseekai2025deepseekr1incentivizingreasoningcapability} and DAPO~\citep{yu2025DAPO} could stabilize training and improve sample efficiency. Parallel efforts like STaR~\citep{Zelikman2022STaR} and MCTS-based reasoning~\citep{hao2023reasoning} promote step-by-step reasoning with minimal supervision. 

\textbf{On the agent side}, systems have evolved from early reactive planning~\citep{yao2022react, xu2023rewoo} to modularized decision pipelines~\citep{liu2023llm+, wu2023autogen}, multi-agent cooperation~\citep{li2023camel, wang2024rethinking}, and embodied interaction~\citep{lin2024advancesembodiednavigationusing, li2025embodiedagentinterfacebenchmarking}. Benchmarks like Sokoban~\citep{sokoban}, FrozenLake~\citep{frozenlake}, and WebShop~\citep{yao2022webshop} provide controlled testbeds for evaluating reasoning under different dynamics. Our work builds on these advances, aiming to unify RL-based reasoning with structured agent training across symbolic and language-centric tasks. We summarize more related works in Appendix~\ref{app:extendedrelatedwork}.
\vspace{-5pt}

\section{Conclusions and Limitations}
\label{sec:conclusions}
We present RAGEN, a general-purpose system for training language agents with reinforcement learning in multi-turn, stochastic environments. Built upon the StarPO framework, RAGEN enables reasoning-guided trajectory optimization and reveals new challenges unique to agent training—such as gradient collapse, rollout drift, and reasoning degradation. Through extensive experiments, we identify key design principles for stabilizing training, including rollout filtering, gradient shaping, and reward-aware reasoning supervision. These insights provide a foundation for building more robust and generalizable LLM agents. Our framework offers a scalable platform for studying autonomous language agents across domains such as symbolic reasoning and web browsing. Limitations of our work include the focus on relatively small-scale tasks, the omission of established RL practices like replay buffers, and the absence of multimodal tasks—which we leave for future work.

\section*{Acknowledgements}

We thank the DeepSeek team for providing the DeepSeek-R1 model and early conceptual inspirations. We are grateful to the veRL team for their infrastructure support, and to the TinyZero team for their discoveries that informed our initial exploration.
We would like to appreciate insightful discussions with Han Liu, Xinyu Xing, Monica Lam, Li Erran Li, John Schulman, Akari Asai, Eiso Kant, Lu Lu, Runxin Xu, Zhihan Liu, Huajian Xin, Zijun Liu, Weiyi Liu, Weimin Wu, Yibo Wen, Jiarui Liu, Lorenzo Xiao, Ishan Mukherjee, Anabella Isaro, Haosen Sun, How-Yeh Wan, Lester Xue, Matthew Khoriaty, Haoxiang Sun, Jiajun Liu.

\bibliographystyle{abbrvnat}
\bibliography{main}
\clearpage
% \onecolumn
\appendix

\section{Background of Reinforcement Learning}
~\label{app:a.rl}
Reinforcement learning (RL) enables foundation models to learn through interaction and reward signals. The general RL objective is:
\begin{equation}
 J(\theta) = \mathbb{E}_{s \sim \mathcal{D}, a \sim \pi_\theta(\cdot|s)}[R(s,a)],   
\end{equation}
where $\pi_\theta$ is the policy, $s$ is the input prompt, $a$ is the response, and $R(s,a)$ is the reward function evaluating response quality. 

Common approaches use reward modeling and policy optimization for RL. Proximal Policy Optimization (PPO) \citep{schulman2017proximalpolicyoptimizationalgorithms} stabilizes training through probability ratio clipping and advantage estimation. The probability ratio is defined as:
\begin{equation}
\rho_t(\theta) = \frac{\pi_\theta(a_t|s_t)}{\pi_{\theta_{old}}(a_t|s_t)}
\end{equation}
The PPO objective uses this ratio with clipping:
\begin{equation}
J_{PPO}(\theta) = \mathbb{E}_t[\min(\rho_iA_i, \hat{\rho_i}A_i) - \beta D_{KL}],
\end{equation}
with probability ratio $\rho_i = \frac{\pi_\theta(o_i|q)}{\pi_{\theta_{old}}(o_i|q)}$ and clipped ratio $\hat{\rho_i} = \text{clip}(\rho_i, 1-\epsilon, 1+\epsilon)$.

For advantage estimation, Generalized Advantage Estimation (GAE) \citep{schulman2018highdimensionalcontinuouscontrolusing} computes:
\begin{equation}
A_t^{GAE(\gamma,\lambda)} = \sum_{l=0}^{\infty}(\gamma\lambda)^l\delta_{t+l}
\end{equation}
where $\delta_t = r_t + \gamma V(s_{t+1}) - V(s_t)$ is the TD error, and $(\gamma,\lambda)$ control the bias-variance tradeoff.

Recently, DeepSeek-R1-Zero~\cite{deepseek1} implements this paradigm through Group Relative Policy Optimization (GRPO), sampling $G$ outputs $\{o_i\}$ [consisting of reasoning and actions] for each prompt and optimizes:
\begin{equation}
J_{GRPO}(\theta) = \mathbb{E}_{q,\{o_i\}}[J_{group}(\theta)],
\end{equation}
where:
\begin{equation}
J_{group}(\theta) = \frac{1}{G}\sum^G_{i=1}\min(\rho_iA_i, \hat{\rho_i}A_i) - \beta D_{KL},
\end{equation}
while mostly similar to Eq. 3, the GRPO advantage is neural-model free and calculated as:
\begin{equation}
A_i = \frac{r_i - \text{mean}(\{r_j\})}{\text{std}(\{r_j\})}.
\end{equation}
Using rule-based rewards $r_i$, this pure RL approach demonstrates emergent reasoning behaviors.

\section{Extended Related Work}
\label{app:extendedrelatedwork}

\noindent\textbf{Reinforcement Learning for Reasoning in LLMs.}~~
Reinforcement learning (RL) on LLMs \citep{christiano2023deepreinforcementlearninghuman,ouyang2022traininglanguagemodelsfollow,chen2021decisiontransformer, havrilla2024teachinglargelanguagemodels} has significantly improved LLMs' reasoning capabilities. Notable approaches include the use of Proximal Policy Optimization Algorithms (PPO) \citep{schulman2017proximalpolicyoptimizationalgorithms} which maintains training stability while enhancing performance by clipping policy updates, Group Relative Policy Optimization (GRPO) \citep{deepseekai2025deepseekr1incentivizingreasoningcapability} for enhancing the ability of systematic problem-solving, actor-critic methods like SAC \citep{haarnoja2018softactorcritic} and ArCHer \citep{archer} that leverages an critic to promote robust exploration and stability, and meta tokens \citep{goyal2024thinkspeaktraininglanguage,herel2024thinkingtokenslanguagemodeling,pfau2024letsthinkdotdot} for structured thinking. Other significant developments include Process Reward Model (PRM) \citep{zhang2025lessonsdevelopingprocessreward, lightman2023verify} and Monte Carlo Tree Search (MCTS) based approaches \citep{hao2023reasoning} for systematic problem-solving. On the other hand, recent advances in LLM reasoning have explored techniques to enable models to generate intermediate chain-of-thought rationales. In particular, STaR \citep{Zelikman2022STaR} iteratively leverages a small set of rationale examples along with a large dataset without rationales. Recent work like SimpleRL-Zoo \citep{zeng2025simplerlzoo} DAPO \citep{yu2025DAPO}, RLOO \citep{rloo}, Dr. GRPO \citep{liu2025DRGRPO}, and Open Reasoner Zero \citep{hu2025openreasonerzero} all demonstrate that minimalist, reproducible RL techniques—featuring decoupled clipping, unbiased optimization, and simple reward schemes—can significantly enhance LLM reasoning performance. 

\noindent\textbf{Existing agent frameworks.}~~
LLM-based agent architectures have evolved from early reasoning-action frameworks \citep{yao2022react, shinn2024reflexion, xu2023rewoo, lin2024swiftsage} to structured approaches \citep{rafa, liu2023llm+, hao2023reasoning, zeng2025reinforcingmultiturnreasoningllm}. Multi-agent systems \citep{du2023improving, li2023camel, chen2023agentverse, wang2024rethinking} are designed for tasks with more complex interactions. Widely used platforms such as OpenAI Gym \citep{brockman2016openaigym} and specialized environments including Sokoban \citep{sokoban}, FrozenLake \citep{frozenlake}, and Webshop \citep{yao2022webshop} provide diverse testbeds for evaluating these agents. Moreover, general-purpose systems \citep{shen2024hugginggpt, wu2023autogen, hao2023toolkengpt, zhuang2023toolchain, xie2023openagents} have enabled broad applications ranging from web navigation and search \citep{qi2025webrltrainingllmweb, searchr1, webagentr1, jin2025empiricalstudyreinforcementlearning}, coding copilot \citep{jimenez2024swebench,deepseekai2024deepseekcoderv2breakingbarrierclosedsource,wang2024mintevaluatingllmsmultiturn} to GUI~\citep{uitars, yao2022webshop}, Game~\citep{hu2025game} and embodied tasks \citep{lin2024advancesembodiednavigationusing, xi2024teachingembodiedreinforcementlearning,li2025embodiedagentinterfacebenchmarking, gigpo}. Social interaction capabilities have been advanced through Generative Agents and AgentSims \citep{park2023generativeagentsinteractivesimulacra,lin2023agentsimsopensourcesandboxlarge}. Challenges persist in architectural complexity and self-correction \citep{he2025enhancingllmreasoning}, especially for diverse, multi-step reasoning tasks \citep{wang2025enigmaevalbenchmarklongmultimodal,nguyen2024turningheatminpsampling, song2024trial}.

\section{Detailed Experimental Settings}
\label{app:exp-setting}

\subsection{Environments and Tasks}
\label{app:env-setting}

\begin{figure}[t]
    \centering
    \includegraphics[width=\linewidth]{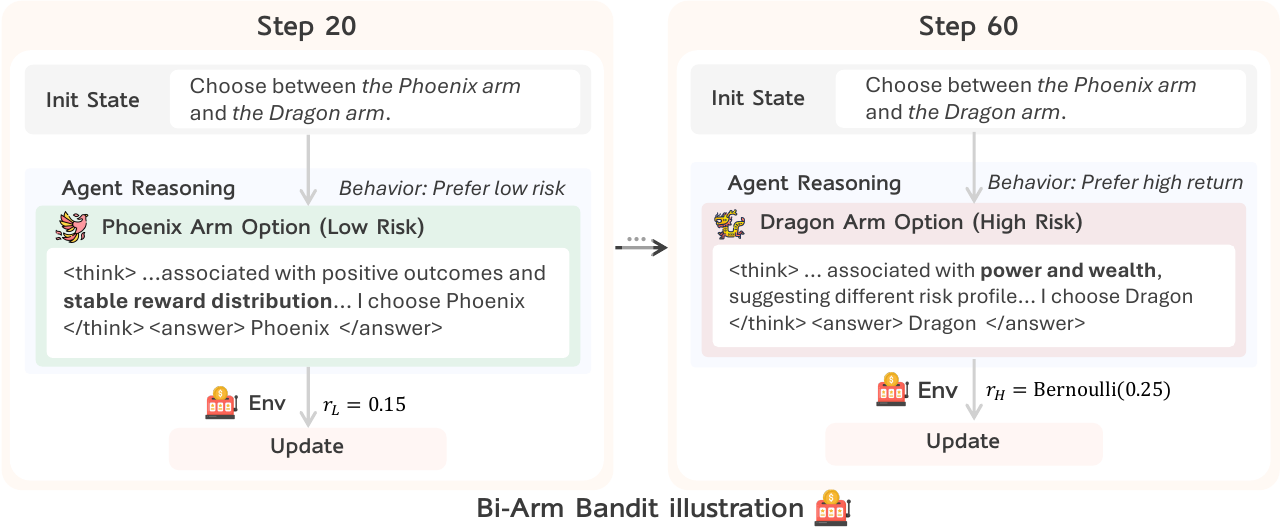}
    \caption{\textbf{Bi-Arm Bandits environment.} The agent chooses between a low-risk arm (Phoenix) and a high-risk yet high-reward arm (Dragon), each linked to symbolic semantics. The agent learns to choose stable reward at early stages and reasons to pursue maximal expected reward and shift toward strategic risk-taking. }
    \label{fig:bandit-intro}
\end{figure}

\begin{figure}[t]
\centering
\begin{minipage}{0.48\textwidth}
\centering
\includegraphics[width=0.8\linewidth]{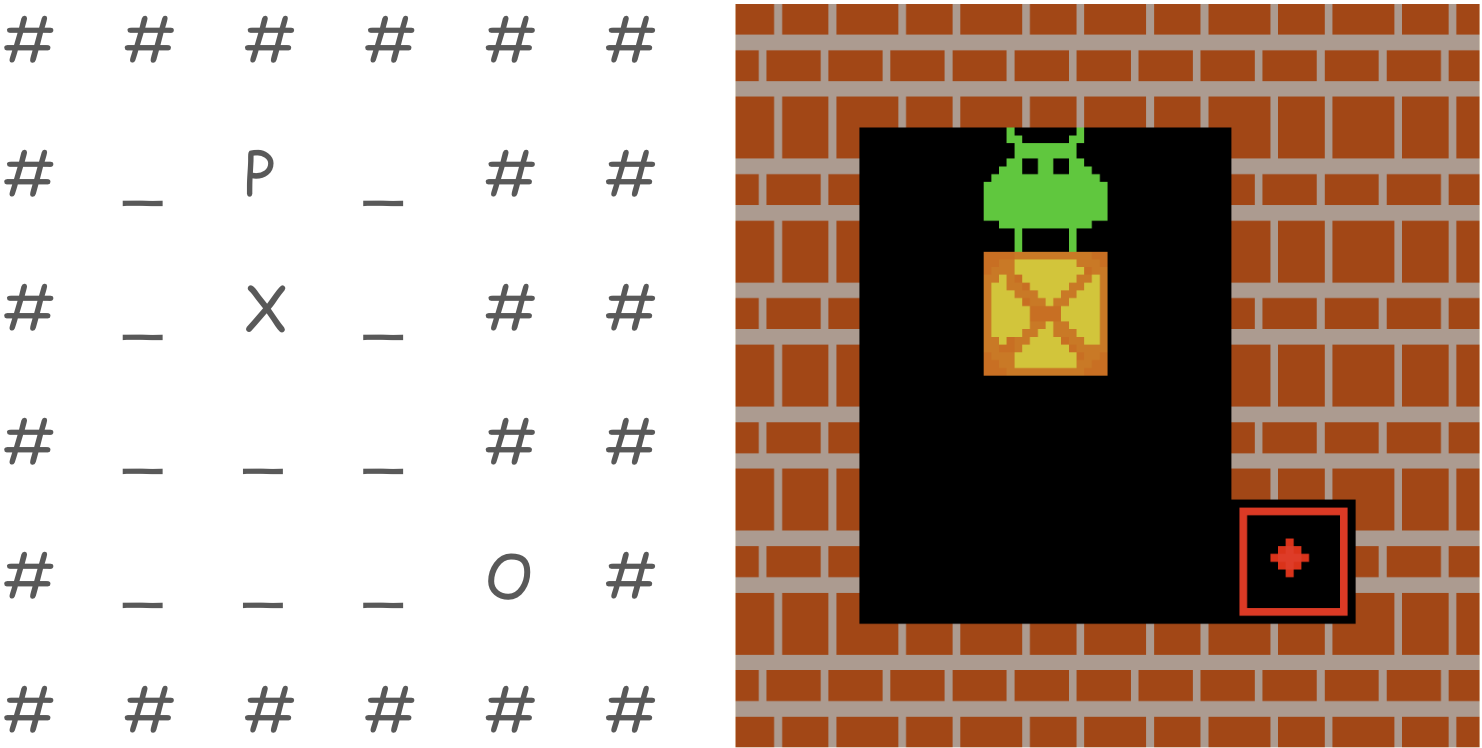}
\subcaption{Sokoban Environment. Actions are irreversible and the boxes cannot be pulled back which needs foresight and careful planning.}
\label{fig:sokoban_env_example}
\end{minipage}
\hfill
\begin{minipage}{0.48\textwidth}
\centering
\includegraphics[width=0.8\linewidth]{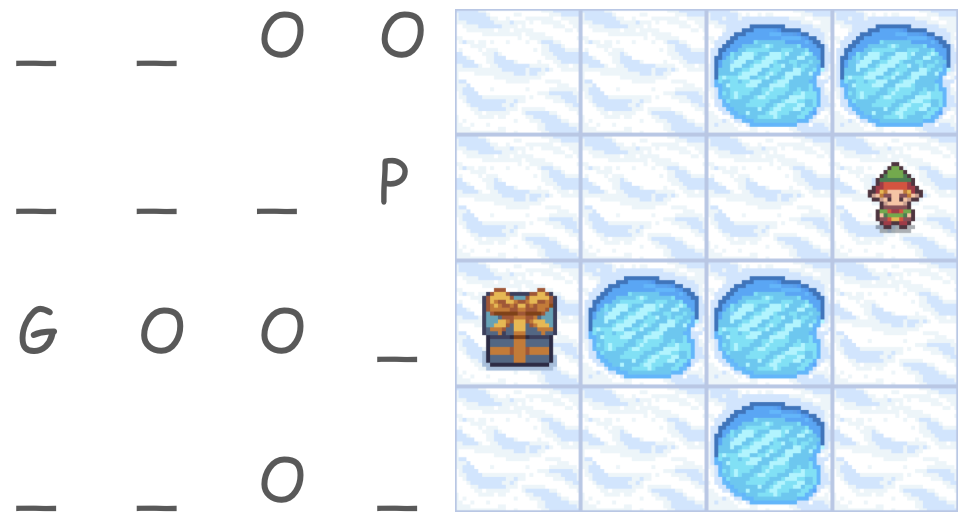}
\subcaption{FrozenLake Environment. Each agent move may randomly slip to a side direction, requiring the agent to adapt under uncertainty.}
\label{fig:frozenlake_env_example}
\end{minipage}
\caption{ \textbf{Sokoban and Frozen Lake environments.}
For each environment, the left shows the agent-observed text rendering; the right is a visual illustration.
(a) Sokoban is a deterministic multi-turn puzzle where the agent pushes boxes onto targets. (b) Frozen Lake combines multi-turn reasoning and stochasticity where the agent needs to reach the gift to succeed.
}
\label{fig:combined_environments}
\end{figure}

We construct a \textbf{diverse four-environment testbed} to evaluate LLM agents across key axes of decision-making complexity. Three environments—Bandit, Sokoban, and Frozen Lake—are symbolic, synthetic, and fully controllable, supporting clean analysis of RL learning from scratch. They are deliberately minimal and decoupled from real-world priors; even large models like GPT-4o perform poorly without training, highlighting the need for grounded policy learning. Complementarily, we include WebShop, a realistic multi-turn task involving natural language grounding and web navigation in a semi-structured interface. Together, the four environments enable systematic study of reasoning, training stability, and generalization in agentic LLMs across symbolic and open-domain settings.

Each environment stresses a distinct capability: Bandits tests reasoning under uncertainty, Sokoban emphasizes irreversible long-horizon planning, Frozen Lake involves stochastic transitions, and WebShop requires language understanding and goal-directed interaction.

\paragraph{Bi-Arm Bandits.}
We design this environment to evaluate whether agents can \textbf{form risk-sensitive hypotheses and revise them based on training}. At each step, the agent must choose between two semantically symbolic options—e.g., “Dragon” vs. “Phoenix”—each linked to a fixed reward distribution (Figure~\ref{fig:bandit-intro}). The low-risk arm always returns a reward of $0.15$, while the high-risk arm samples from $Bernoulli(0.25)$: higher variance, higher expected return. 

Importantly, the low-risk arm wins more often per trial, even though the high-risk arm is better in expectation. This designed to test reasoning: without inductive bias, models may prefer the lo-arm due to its more frequent success, but a reasoning agent must learn to associate symbolic cues (e.g., "Dragon") with underlying reward statistics, override misleading short-term signals, and “justify” high-risk choices based on long-term expected return. We further test this by reversing the symbolic labels to probe agent’s reasoning under opposed reward systems.

\noindent\textbf{Sokoban.}~~~
We use the puzzle Sokoban (Figure~\ref{fig:sokoban_env_example}) to study multi-turn agent interaction. The agent must push a box to the goal in a grid within constrained steps. Unlike standard navigation, Sokoban is irreversible: boxes can only be pushed, not pulled back, which requires the agent to reason ahead to avoid dead-ends. The reward signal encourages efficiency and accuracy: $+1$ for each box on target, $-1$ for off-target boxes, $+10$ upon task completion, and $-0.1$ per action.

\noindent\textbf{Frozen Lake.}~~~
This environment (Figure~\ref{fig:frozenlake_env_example}) combines long-horizon decision-making with stochastic transitions. The agent navigates a grid with slippery tiles; each action succeeds with probability $1/3$ and deviates perpendicularly with probability $2/3$. The agent should reach the goal without falling into holes. Rewards are sparse: successful trials receive a reward of $+1$, with all others $0$. 

\noindent\textbf{WebShop.}~~~
To complement the symbolic environments, we include WebShop~\cite{yao2022webshop}, a multi-turn web-based shopping task that tests agents' ability to ground natural language queries, navigate semi-structured interfaces, and retrieve goal-relevant information. The agent must select a product that matches a user’s request by issuing search queries, clicking links, and reading product descriptions—introducing realistic language grounding and action space challenges absent in purely symbolic settings.

\subsection{Training and Evaluation Settings}
\label{app:train-eval-setting}
We conduct our experiments using Qwen2.5-0.5B-Instruct~\citep{qwen2.5}, trained via the StarPO variants with a maximum of 200 rollout–update iterations on NVIDIA H100/A100 GPUs leveraging the veRL~\footnote{https://github.com/volcengine/verl} repository. Considering the extremely long training time of WebShop due to its long-context nature, we train on WebShop for 100 steps. Each rollout consists of $K=16$ trajectories per environment group, based on prompt size $P=8$ and maximum 5 interaction turns per episode. Agents are allowed up to 5 actions per turn and 10 actions per episode. The update batch size is $E=32$, with mini-batch size 4 per GPU. Policy optimization uses GAE with $(\gamma=1.0, \lambda=1.0)$ and Adam with $(\beta_1, \beta_2) = (0.9, 0.999)$.
We use entropy regularization ($\beta=0.001$) For experiments with vanilla StarPO we use a KL coefficient of 0.001, using the $\rm k1$ estimation\footnote{http://joschu.net/blog/kl-approx.html}. without KL loss term during training, following~\citep{yu2025DAPO}, and track KL post-hoc. We impose a format penalty of $-0.1$ if the agent fails to output valid structured responses (e.g., missing \texttt{<think>} or \texttt{<answer>} tags), encouraging adherence to response conventions. To accelerate rollout generation, we disable enforce\_eager and retain the computation graph across prefill and sampling in vLLM. We utilize Fully Sharded Data Parallel (FSDP) training strategy for multi-GPU experiments. For distributed training, we employ Ray as the multi-processing backend with XFORMERS attention implementation.

For evaluation, we choose a fixed 256 input prompts per environment and decode using temperature $T{=}0.5$, sampling stochastically to better capture robustness in agent behaviors. Episode truncation occurs after 5 turns or 10 total actions.

\subsection{Evaluation Metrics}
\label{app:metric-setting}

To track agent learning dynamics and detect training instabilities, we monitor the following metrics throughout training. Except for the success rate, which is evaluated on a fixed validation set, all metrics are computed over validation instances.

\begin{itemize}
    \item \textbf{Average Success Rate.} Measures task completion accuracy on a fixed set of validation prompts. An episode is considered successful if the agent solves the task (e.g., pulling the high-reward arm in Bandit, pushing all boxes to targets in Sokoban, reaching the goal in Frozen Lake, and a successful purchase in WebShop).

    \item \textbf{Rollout Entropy.} Computes the average token-level entropy of sampled responses, capturing the exploration level and policy uncertainty. A sharp entropy drop may indicate premature policy convergence or collapse.

    % \item \textbf{KL Divergence.} We track two forms of KL: (1) divergence between the current policy and the initial (pretrained) model, reflecting deviation from prior behavior, and (2) divergence between consecutive policy checkpoints, serving as a stability indicator for policy updates.

    \item \textbf{In-Group Reward Variance.} Measures reward standard deviation across rollouts sampled from the same prompt group. High in-group variance reflects diverse behaviors and learning potential; a sudden collapse indicates reward homogenization and policy stagnation.

    \item \textbf{Total Response Length.} Average number of tokens generated per rollout, measuring the verbosity and reasoning depth of the agent. Fluctuations in length may signal changes in planning style or confidence.

    \item \textbf{Gradient Norm.} $\ell_2$ norm of the policy gradient vector, used as a proxy for training stability. Spikes often correlate with phase transitions in policy behavior or unstable reward signals.

\end{itemize}

These metrics provide complementary views of policy quality, update dynamics, and reasoning behavior, helping diagnose when and why agent training succeeds or fails.

\section{Results on Larger Models and Various Optimization Algorithms}
\label{app:largermodelabaltion}
We extend all our evaluation to 3B/7B/72B scale model and explore the effects of various algorithm choices like KL removal and asymmetric clipping, together with turn-aware optimization techniques including Generalized Advantage Estimation (GAE) and response masking.

\begin{figure}[t]
    \centering
    \includegraphics[width=0.95\linewidth]{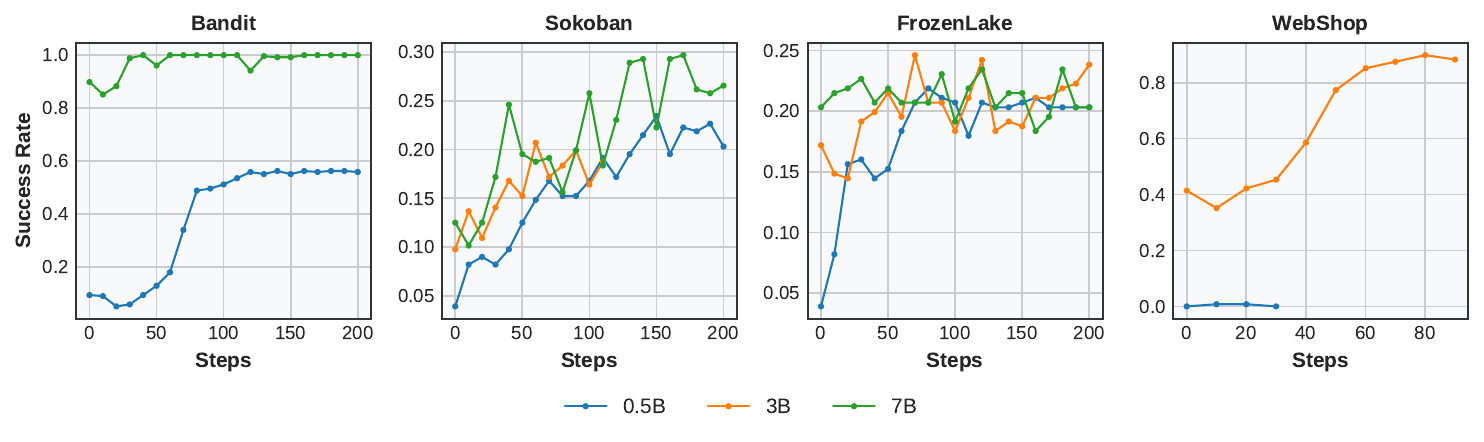}
    \caption{\textbf{Scaling effect across environments.} The larger model outperforms smaller models on Bandit and WebShop, which benefit from language priors, but shows limited gains on symbolic, grid-based environments like Sokoban and FrozenLake.}
    \label{fig:scaling-effect}
\end{figure}

\paragraph{Scaling Effects.} We extend our trained model to 3B / 7B to evaluate scaling effects of RL training. Results can be found in Figure~\ref{fig:scaling-effect}. Note that due to the extremely long context length of WebShop could due to OOM Error within 4xH100 for 7B models, we only report 3B performance in WebShop task.  On \textbf{Bandit} and \textbf{WebShop}, the larger model demonstrates significantly stronger performance compared to smaller models. However, the improvements are marginal on \textbf{Sokoban} and \textbf{FrozenLake}. We attribute this discrepancy to the nature of the environments: Sokoban and FrozenLake are symbolic, grid-based tasks with minimal overlap with pretraining data, limiting the model’s ability to leverage language priors. In contrast, Bandit and WebShop involve natural language interactions, allowing pretrained models to more effectively exploit linguistic patterns for policy learning, even in the absence of explicit environment dynamics. This can be further validated through cases in Figure~\ref{fig:bandit-multiscale},~\ref{fig:sokoban-multiscale},~\ref{fig:frozenlake-multiscale},~\ref{fig:webshop-multiscale}, where semantic-rich tasks like Bandit and WebShop presents significantly more diverse reasoning patterns and better benefit from scale. 

\paragraph{Frontier Model Performance.} To contextualize our small model’s performance, we evaluate two large foundation models—\textbf{GPT-4o} and \textbf{Qwen2.5-72B-Instruct}—on \texttt{SimpleSokoban} and \texttt{FrozenLake} in a zero-shot setting. Both models are prompted with task instructions and example formats, without any fine-tuning or in-context trajectory rollouts. See Table~\ref{tab:foundation-models}

\begin{table}[t]
\centering
\small
\begin{tabular}{lcccc}
\toprule
\textbf{Model} & \textbf{Sokoban (\%)} & \textbf{FrozenLake (\%)} & \textbf{Response Length} & \textbf{Effective Action (\%)} \\
\midrule
GPT-4o               & \textbf{27.73} & \textbf{26.56} & 190.3 / 132.1 & 70.0 / 88.7 \\
Qwen2.5-72B          & 19.53 & 23.83 & 179.4 / 107.3 & 67.8 / 87.3 \\
\midrule
\textbf{0.5B (ours)} & 20.70 & 21.48 & 89.5 / 110.6 & \textbf{79.2} / 80.7 \\
\bottomrule
\end{tabular}
\vspace{2pt}
\caption{\textbf{Zero-shot vs. trained performance.} Our 0.5B model, trained with only 4 responses per prompt, achieves performance comparable to that of large foundation models without any fine-tuning. Response length and effectiveness values for GPT-4o / Qwen are reported per environment.}
\label{tab:foundation-models}
\end{table}

While GPT-4o and Qwen2.5-72B achieve 19–28\% success rates in Sokoban and FrozenLake without any task-specific adaptation, our 0.5B model reaches \textbf{20.70\%} and \textbf{21.48\%} respectively after training from scratch. This result is notable given that our model has over \textbf{100$\times$ fewer parameters}. It highlights that even under strict resource constraints, careful rollout construction and policy optimization (see Section~\ref{sec:rollout}) can match the generalization ability of significantly larger models.

\begin{figure}[t]
  \centering
  \includegraphics[width=0.7\textwidth]{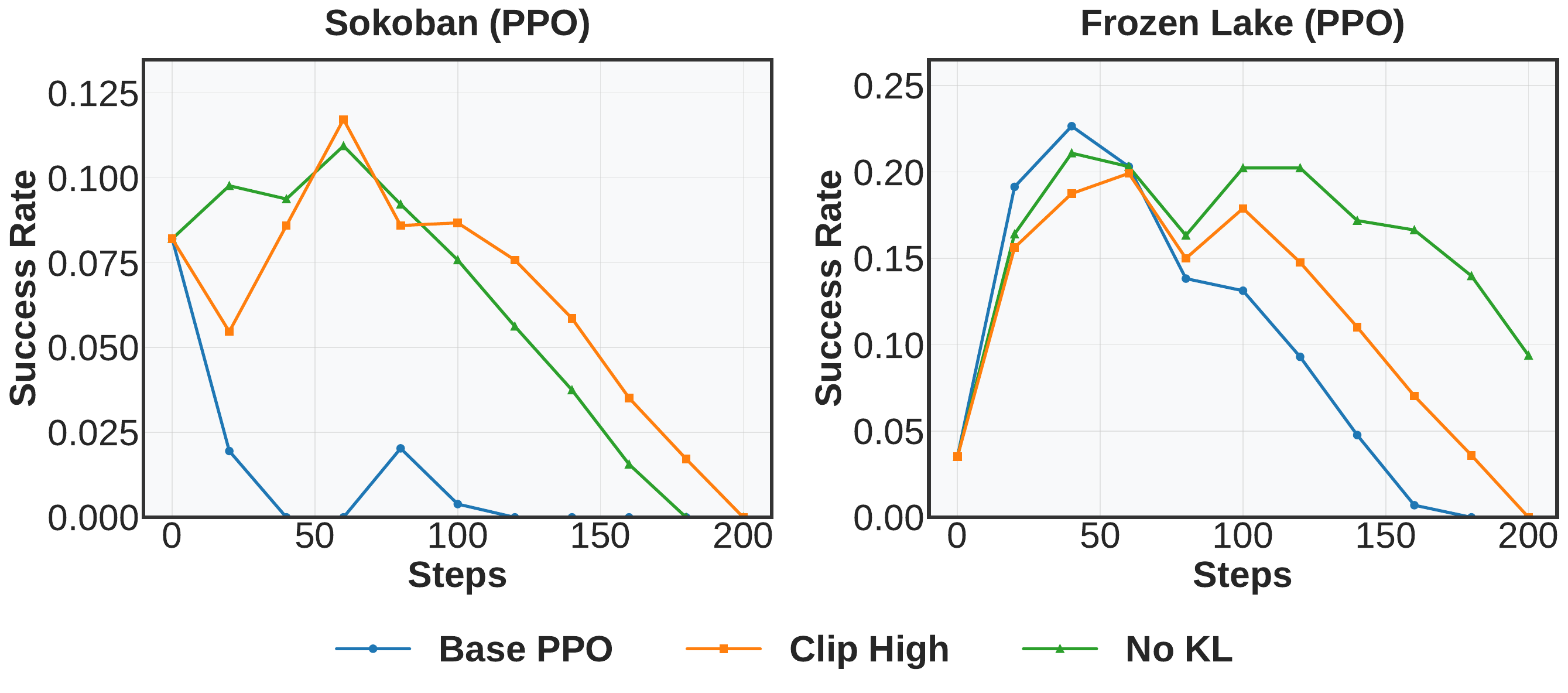}
  \caption{
    \textbf{Effect of KL removal and asymmetric clipping on PPO stability.}
    The two designs both improve peak performance and delay collapse in multi-turn RL.
  }
  \label{fig:trick_results}
\end{figure}

\paragraph{Gradient Shaping.} We evaluate the effectiveness of KL Term Removal and Clip-Higher~\cite{yu2025DAPO}, finding them useful by simply extending from single-turn static tasks to agent tasks:
\begin{itemize}[left=2pt, itemsep=2pt, topsep=4pt]
    \item \textbf{KL Term Removal:} We eliminate the KL divergence penalty from PPO’s objective, relying only on policy loss and entropy bonus for gradient updates. It removes the constraint to stay close to the initial model distribution and encourage the model to explore.
    
    \item \textbf{Clip-Higher (Asymmetric Clipping):} We decouple the PPO clipping range by using a higher upper bound ($\epsilon_\text{high} = 0.28$) than the lower bound ($\epsilon_\text{low} = 0.2$). It allows the model to learn more aggressively from high-reward rollouts for more effective training.
\end{itemize}
\vspace{-5pt}
As shown in Figure~\ref{fig:trick_results}, both methods boost the success rate and extend stable training phases, showing how multi-turn RL benefits from more flexible gradient shaping.

\begin{figure}[t]
    \centering
    \includegraphics[width=\linewidth]{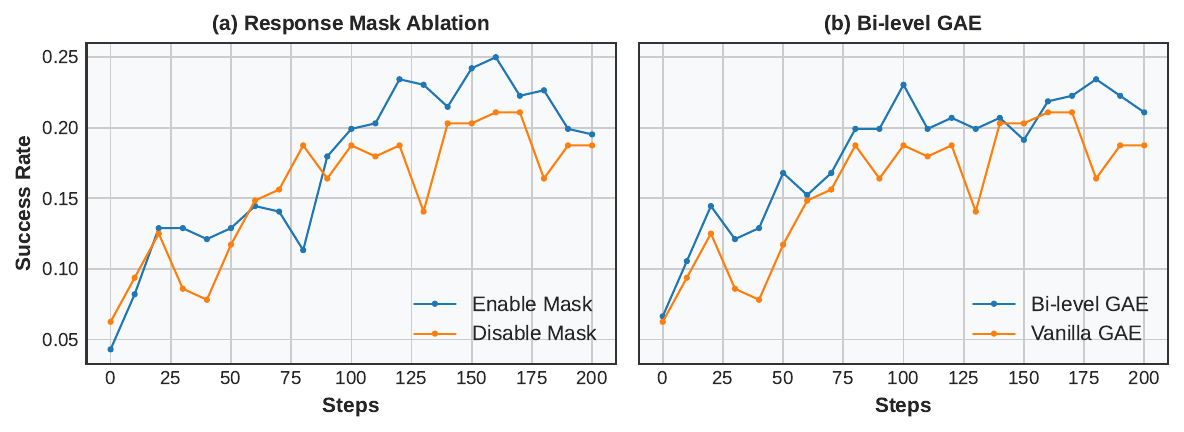}
    \caption{\textbf{Ablation on turn-aware optimization strategies on Sokoban Task}. Both response masking and bi-level GAE improve multi-turn RL performance.}
    \label{fig:turn-aware-ablation}
\end{figure}

\paragraph{Response Masking and Bi-Level GAE.}
Following the turn-aware optimization strategy proposed in~\cite{vagen}, we evaluate the effect of response masking and bi-level GAE on a 0.5B model. As shown in Figure~\ref{fig:turn-aware-ablation}, both techniques contribute to improved performance in multi-turn RL tasks, which highlight the promise of turn-aware RL training algorithms  for stabilizing and enhancing language agent training.

\section{When Does Uncertainty-Based Filtering Help?}
\label{app:std}

We hypothesize that the effectiveness of StarPO-S largely depends on the variance of rollout rewards within each environment. In settings where the task is either too easy or too difficult, the generated trajectories tend to have very low intra-group variance—meaning the model is overconfident or uniformly poor across samples. In such cases, standard StarPO may propagate misleading gradients, while StarPO-S helps by filtering out low-confidence rollouts. Conversely, in open-ended or more diverse environments (like WebShop), the rollout variance tends to be naturally high, which reduces the marginal benefit of StarPO-S filtering.

\begin{figure}[t]
    \centering
    \includegraphics[width=\linewidth]{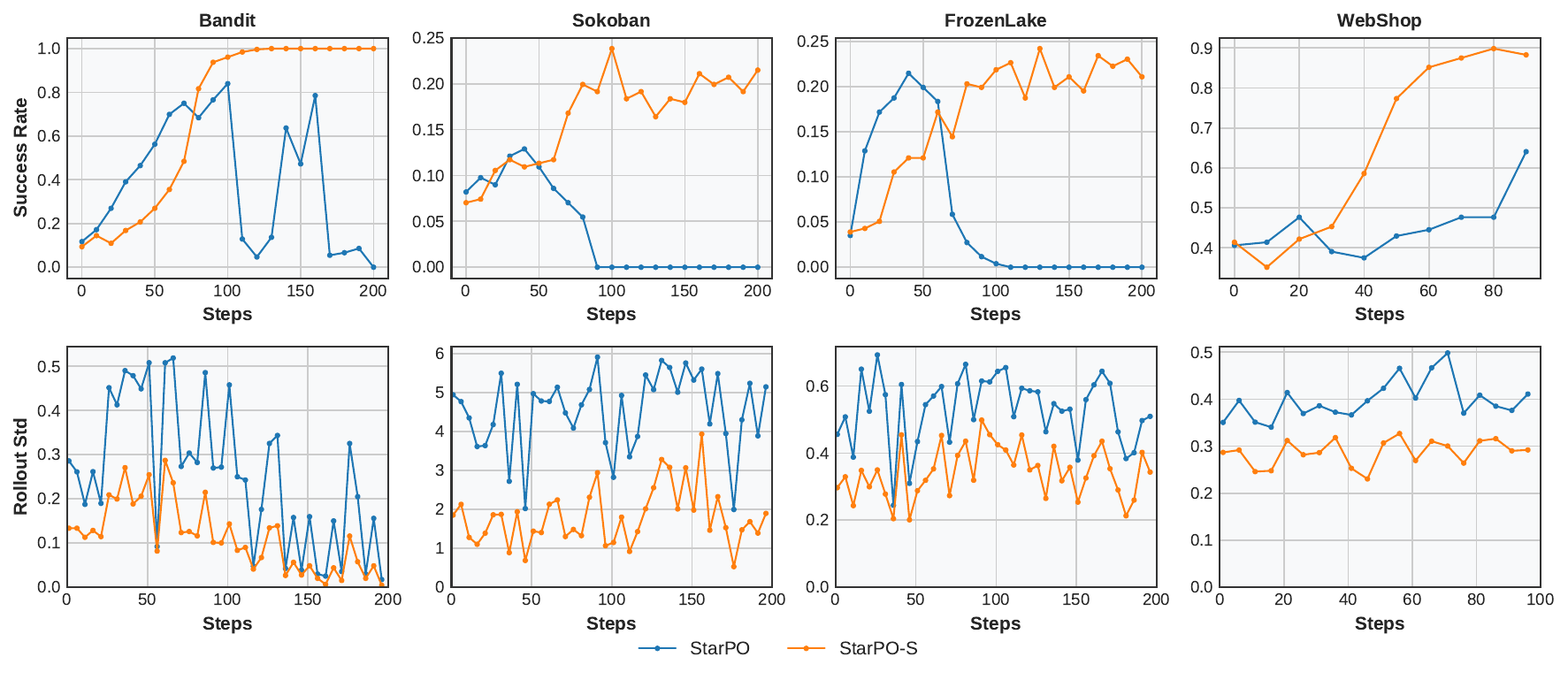}
    \caption{\textbf{Comparison of success rate (top) and rollout variance (bottom)}. StarPO-S basically improve training stability on the environments having extremely easy or difficult problems like Sokoban and Frozen Lake. This would ead to instances with small rollout Stds which can be easily filtered out through StarPO-S to make training more stable. On Tasks like WebShop, the rollout Std is already consistently high, and StarPO itself already can achieve good performance. }
    \label{fig:starpo_s_rollout_std}
\end{figure}

Figure~\ref{fig:starpo_s_rollout_std} supports this intuition. The top row shows the success rates of StarPO and StarPO-S across four environments, while the bottom row shows the evolution of \texttt{in\_group\_std} and \texttt{chosen\_in\_group\_std} over training. In Bandit, Sokoban, and FrozenLake, StarPO-S consistently outperforms StarPO, with the gap widening as the rollout variance drops. In WebShop, however, the high and stable variance suggests more diversity in generated responses, making the filtering effect of StarPO-S less critical—explaining the smaller performance gap.

These results indicate that StarPO-S is most beneficial when environments exhibit low rollout uncertainty, providing a simple diagnostic for when to apply it.

\section{Case Study: The Emergence of Echo Trap with RL}
\label{app:echotrap}
\begin{table}[t]
\centering
\begin{tabular}{@{}p{0.07\textwidth} L{0.9\textwidth}@{}}
\toprule
\textbf{Step} & \textbf{Model Output} \\
\midrule
\rowcolor{gray!10}
0 & Dragon is often associated with patience, strength, and reliability, while Phoenix is associated with speed, agility, and creativity. \\
\rowcolor{gray!10}
0 & The symbolic meaning of Dragon and Phoenix's names suggests that they may have different reward distributions... Therefore, I think Phoenix is more likely to give higher rewards on average. \\
\rowcolor{red!5}
150 & My initial intention is to pick arm Dragon. My first action is to go for arm Dragon. \\
\rowcolor{red!5}
150 & My initial intention is to pick arm Dragon, so my first action is to go for arm Dragon. \\
\rowcolor{red!5}
150 & My initial intention is to choose Arm Dragon. After analyzing both arms, my intention was to choose Arm Dragon. \\
\bottomrule
\end{tabular}
\vspace{1pt}
\caption{\textbf{Examples of reasoning patterns in the Bandit task.} Top rows show diverse reasoning from model before training; bottom rows show repetitive and collapsed reasoning after RL training.}
\vspace{-3pt}
\label{tab:bandit_reasoning_drift}
\end{table}

\begin{figure}[t]
  \centering
 \includegraphics[width=\textwidth]{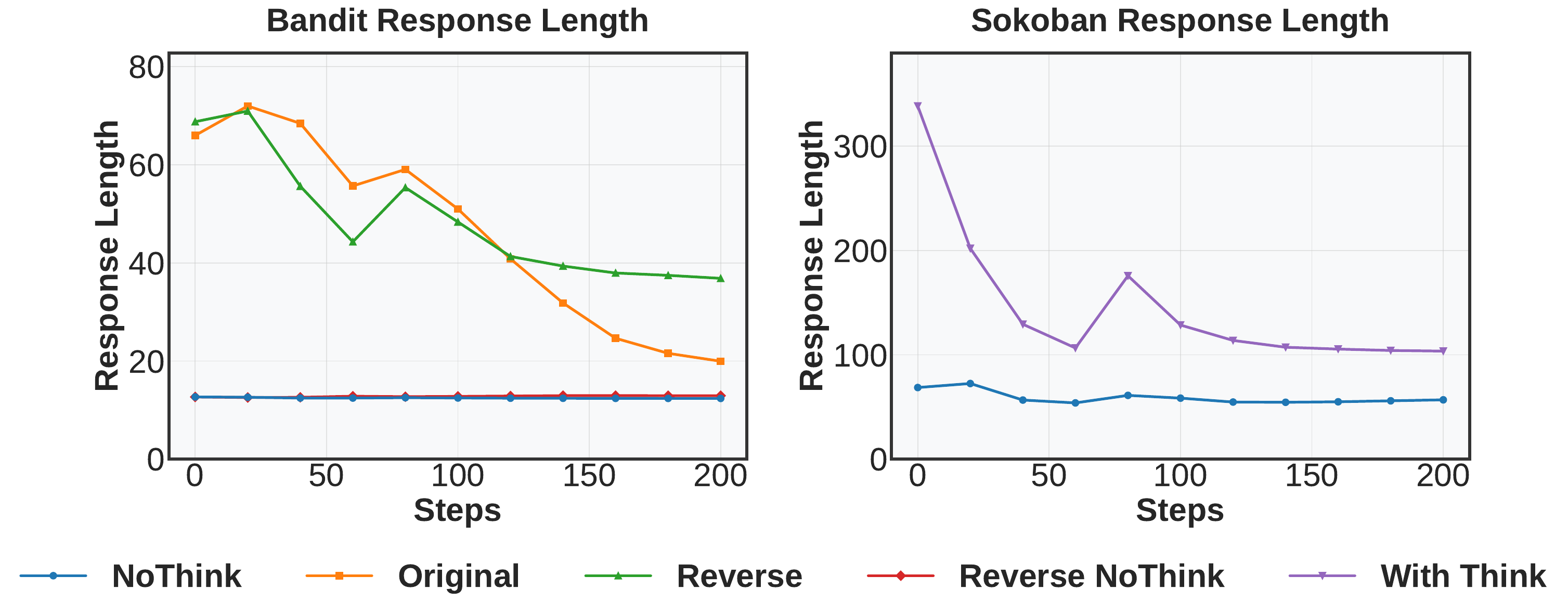}
\caption{\textbf{Reasoning length over training iterations across different tasks.} We track the average token count of reasoning segments (\texttt{<think>} blocks) during RL training. Across all environments, reasoning length declines as training progresses, with \texttt{BanditRev} maintaining longer traces—possibly due to greater semantic-reward conflict requiring more deliberation.}

  \label{fig:reasoning_decay}
\end{figure}

We show cases that demonstrates the Echo Trap in RL training. As shown in Table~\ref{tab:bandit_reasoning_drift}, the top rows (Step 0) display varied hypotheses about Dragon and Phoenix, while the bottom rows (Step 150) converge to near-identical phrasing focused on “choosing Dragon” without justification.

\section{Comparing Agent RL with Supervised Fine-Tuning}
% We implement a "no-think" training variant, SarPO (State-Action-Reward Policy Optimization), where the model's intermediate reasoning steps are masked, preserving only the final answer enclosed in <answer> tags. This approach focuses the learning process on direct state-to-action mapping rather than explicit reasoning chains. The configuration supports both standard RL training with complete reasoning chains and SarPO variants through a boolean flag (enable\_think), allowing for comparative analysis of the impact of reasoning on policy learning.

Apart from StarPO for RL training, we also employ Supervised Fine-tuning (SFT) as another agent training approach, evaluating it on the Sokoban and Frozen Lake task.
We employ LoRA with a rank of 64 and an alpha value of 32, targeting all linear layers in the model. The SFT process uses a learning rate of 1e-4 with a training batch size of 128. 
We generate ground-truth trajectory data through breadth-first search (BFS), setting a maximum depth of 100 to create 1,000 training samples and 100 test samples. For SFT, we structure the multi-turn interaction as a conversational format. At each turn, the model must generate the next action from the ground-truth trajectory, encapsulating its response within \texttt{<answer> </answer>} tags to maintain format consistency.
% To ensure efficient training, we implement gradient checkpointing and set the maximum sequence length to 2,048 tokens.

We analyze the comparative performance of SFT against our stable RL baseline StarPO-S. SFT achieves 74.6\% and 23\% performance on Sokoban and Frozen Lake, respectively, Compared to the 20.3\% and 21.8\% performance with StarPO-S. The results indicate that SFT demonstrates superior performance to RL approaches. We draw conclusions from the results that although rule-based RL show promise for agent tasks, there is still a need to build more scalable and effective agent RL algorithms to achieve human-comparable performance with solely model self-evolution.
\section{Efficient Training with Low-Rank Adaptation (LoRA)}
\paragraph{Motivation.}
While the main body of the paper reports results obtained by full‑parameter fine–tuning, in practice such a setting may be prohibitive when scaling to larger models or longer‐horizon tasks.  We therefore implement a parameter‑efficient variant of RAGEN based on Low‑Rank Adaptation~\citep{hu2021lora}.\footnote{We set rank~$r{=}64$, $\alpha{=}64$, and inject adapters into all linear projections of the transformer blocks. We also increased learning rate by $10\times$ for both actor and critic.}

\noindent\textbf{Performance parity.}~~
Despite updating only a fraction of the model parameters, LoRA reaches a validation success rate comparable to that achieved by full fine‑tuning of the entire network for the SimpleSokoban task, achieving approximately a $0.2\%$ success rate on the validation set.

\noindent\textbf{Resource savings.}~~
We compare the hardware footprint of LoRA with full fine‑tuning.  Across an 80‑minute training horizon we measure:

\begin{itemize}[nosep,leftmargin=*]
    \item \textbf{GPU memory.}  LoRA stabilizes at $\mathbf{\approx 23\%}$ of device memory versus $\mathbf{\approx 48\%}$ for full updates, cutting the peak allocation by \textgreater50\,\%.
    \item \textbf{GPU utilization.}  Average GPU utilization drops from \(\sim\!34\%\) to \(\sim\!14\%\).
    \item \textbf{Power consumption.}  Mean power draw decreases from \(\sim\!22\%\) to \(\sim\!12\%\), a \(\approx45\%\) reduction.
\end{itemize}

% These gains translate into lower cost per training step and enable experimentation on commodity hardware without sacrificing performance.

\noindent\textbf{Take‑aways.}~~
Parameter‑efficient fine‑tuning provides a practically viable alternative for RAGEN: it attains comparable policy quality while more than halving memory, compute, and power demands.  Consequently, future work that scales StarPO to larger backbones or longer contexts can adopt LoRA (or other adapter‑based methods) as the default optimization strategy without re‑engineering the training loop.

\section{PPO Failure Mode in Frozen Lake}
\label{app:frozenlake-grpo}
Among the three evaluated environments, we observe an interesting divergence on Frozen Lake: PPO tends to collapse earlier or converge less stably than GRPO. This contrasts with the general trend where PPO demonstrates better performance, prompting further analysis.

One possible explanation lies in the environment’s long-horizon stochasticity. In Frozen Lake, agent actions always lead to highly non-deterministic transitions, and intermediate states can appear similar while leading to very different outcomes. This makes value estimation challenging. As PPO relies on a learned value function, instability in critic learning may amplify optimization noise and contribute to early collapse.
GRPO, by contrast, does not rely on explicit value learning. Its reward-weighted update procedure may be more tolerant to uncertainty in these settings, leading to comparatively more stable training on Frozen Lake—even if it remains less effective in other tasks. Overall, we summarize environments with high stochasticity may pose greater challenges for value-based methods, and that critic-free approaches can serve as a useful baseline in such cases.

\section{Prompt Templates}
\subsection{Bi-Arm Bandit Environment Prompts}
The Bi-Arm bandit environment implements a classic reinforcement learning problem where an agent must balance exploration and exploitation. We present the prompt templates below.
\begin{tcolorbox}[colback=gray!2, colframe=black, boxrule=0.5pt, arc=2pt, width=0.95\linewidth, title=Model Templates]

\texttt{<|im\_start|>[system]:\\
\{prompt\}\\
You're a helpful assistant. You always respond by giving your answer in <answer>...</answer>. Max response length: 200 words (tokens).\\
<|im\_end|>\\
<|im\_start|>[user]:\\
\{prompt\}\\
You are playing a bandit game. Goal: Maximize your total reward by choosing which arm to pull. \\
Game Rules: \\
1. There are 2 arms, named {name\_a} and {name\_b}\\
2. Each arm has its own reward distribution, related to their names. \\
3. Analyze the symbolic meaning of each arm's name to guess how their reward distribution might behave.\\
4. Based on the symbolic meaning of their names, which arm do you think is more likely to give higher rewards on average? Choose between {name\_a} and {name\_b}, and output like <answer> {name\_a} </answer> or <answer> {name\_b} </answer>.\\
<|im\_end|>\\
<|im\_start|>assistant\\
<think>}

\end{tcolorbox}

\subsection{Sokoban Environment Prompts}

The Sokoban environment presents a classic puzzle game where an agent must push boxes to target locations. The following sections detail the prompt structure used to interface with language models.

\begin{tcolorbox}[colback=gray!2, colframe=black, boxrule=0.5pt, arc=2pt, width=0.95\linewidth, title=Model Templates]
\texttt{<|im\_start|>system\\
\{prompt\}\\
You're a helpful assistant. You always respond by first wrapping your thoughts in <think>...</think>, then giving your answer in <answer>...</answer>. Max response length: 200 words (tokens).\\
<|im\_end|>\\
<|im\_start|>user\\
\{prompt\}\\
You are solving the Sokoban puzzle. You are the player and you need to push all boxes to targets. When you are right next to a box, you can push it by moving in the same direction. You cannot push a box through a wall, and you cannot pull a box. The answer should be a sequence of actions, like <answer>Right || Right || Up</answer>\\
<|im\_end|>\\
<|im\_start|>assistant\\
<think>}
% \textbf{Base Model Template:}
% \texttt{A conversation between User and Assistant. The user asks a question, and the Assistant solves it. The assistant first thinks briefly about the reasoning process in the mind and then provides the user with the answer.\\
% User: \{prompt\}\\
% Show your work in <think> </think> tags. And return the final answer in <answer> </answer> tags, for example <think> [Thoughts] </think> <answer> 1 </answer>\\
% Assistant:\\
% <think>}
\end{tcolorbox}

% The task introduction provides comprehensive information about the game mechanics, symbols, and reward structure:

% \begin{promptbox}{Task Introduction}
% You are a Sokoban solver.

% \textbf{Sokoban Quick Guide}\\
% Goal: Push all boxes (X) onto targets (O).

% \textbf{Symbols:}\\
% \# Wall\\
% \_ Floor\\
% O Target\\
% X Box\\
% P You\\
% $\checkmark$ Box on Target\\
% S You on Target

% \textbf{Rules:}\\
% Push boxes (can't pull).\\
% Avoid walls (\#).

% \textbf{Answers:}\\
% \texttt{<answer> Up </answer>}\\
% \texttt{<answer> Down </answer>}\\
% \texttt{<answer> Left </answer>}\\
% \texttt{<answer> Right </answer>}

% \textbf{Rewards:}\\
% Move: -0.1\\
% Box on target: +1.0\\
% All boxes placed: +10.0
% \end{promptbox}

The environment uses a grid-based representation with specific symbols for different elements:

\begin{tcolorbox}[colback=gray!2, colframe=black, boxrule=0.5pt, arc=2pt, width=0.95\linewidth, title=Grid Representation]
\texttt{
The meaning of each symbol in the state is:\\
\#: wall, \_: empty, O: target, $\checkmark$: box on target, X: box, P: player, S: player on target\\
}
\end{tcolorbox}

% \begin{figure}[H]
%     \centering
%     \includegraphics[width=\linewidth]{figures/sokoban_env_example.png}
%     \caption{Grid representation and rendered image of an example Sokoban Environment}
%     \label{fig:sokoban_env_example}
% \end{figure}

% \begin{figure}[H]
%     \centering
%     \includegraphics[width=\linewidth]{figures/frozenlake_env_example.png}
%     \caption{Grid representation and rendered image of an example FrozenLake Environment}
%     \label{fig:frozenlake_env_example}
% \end{figure}

The instruction template only consists of available actions and restrictions:

\begin{tcolorbox}[colback=gray!2, colframe=black, boxrule=0.5pt, arc=2pt, width=0.95\linewidth, title=Instruction Template]
\texttt{Your available actions are:\\
Up, Down, Left, Right\\
You can make up to 10 actions, separated by the action separator " || "
}
\end{tcolorbox}

% The environment provides state updates through different rendering modes:

% \begin{promptbox}{State Rendering Modes}
% \textbf{Available Modes:}\\
% tiny\_rgb\_array: Text-based grid representation\\
% list: Space-separated grid symbols\\
% state: Numerical state matrix\\
% rgb\_array: Visual representation as image

% \textbf{Rendering Format:}\\
% Grid cells are separated by tabs horizontally\\
% Rows are separated by newlines\\
% Each cell is padded with spaces for readability
% \end{promptbox}

\subsection{FrozenLake Environment Prompts}

The FrozenLake environment implements a grid-world navigation task where an agent must traverse a slippery frozen surface to reach a goal. Below we detail the prompt structure used for this environment.

\begin{tcolorbox}[colback=gray!2, colframe=black, boxrule=0.5pt, arc=2pt, width=0.95\linewidth, title=Model Templates]
% \textbf{Instruct Model Template:}

\texttt{<|im\_start|>system\\
\{prompt\}\\
You're a helpful assistant. You always respond by first wrapping your thoughts in <think>...</think>, then giving your answer in <answer>...</answer>. Max response length: 200 words (tokens).\\
<|im\_end|>\\
<|im\_start|>user\\
\{prompt\}\\
You are solving the FrozenLake puzzle. Forbid the whole and go to the target. You may move to the unintended direction due to the slippery ice. Example answer format: <think>To forbid the hole and go to the target, I should go left then go up.</think><answer>Left || Up</answer>\\
<|im\_end|>\\
<|im\_start|>assistant\\
<think>}

% \textbf{Base Model Template:}

% \texttt{A conversation between User and Assistant. The user asks a question, and the Assistant solves it. The assistant first thinks briefly about the reasoning process in the mind and then provides the user with the answer.\\
% User: \{prompt\}\\
% Show your work in <think> </think> tags. And return the final answer in <answer> </answer> tags, for example <think> [Thoughts] </think> <answer> 1 </answer>\\
% Assistant:\\
% <think>}
\end{tcolorbox}

% \begin{promptbox}{Task Introduction}
% You are walking on a frozen lake.

% \textbf{FrozenLake Quick Guide}\\
% Goal: Reach the goal (G).

% \textbf{Symbols:}\\
% \_ Frozen\\
% O Hole\\
% G Goal\\
% P Player

% \textbf{Rules:}\\
% Avoid falling into holes (O).\\
% Frozen tiles are slippery, you may move perpendicular to your intended direction.

% \textbf{Answers:}\\
% \texttt{<answer> Up </answer>}\\
% \texttt{<answer> Down </answer>}\\
% \texttt{<answer> Left </answer>}\\
% \texttt{<answer> Right </answer>}

% \textbf{Rewards:}\\
% Fall into hole: 0\\
% Reach goal: +1.0
% \end{promptbox}

The environment uses a grid-based representation with specific symbols for different elements:

\begin{tcolorbox}[colback=gray!2, colframe=black, boxrule=0.5pt, arc=2pt, width=0.95\linewidth, title=Grid Representation]
\texttt{
The meaning of each symbol in the state is:\\
P: player, \_: empty, O: hole, G: goal, X: player in hole, $\checkmark$: player on goal
}
\end{tcolorbox}

The instruction template only consists of available actions and restrictions:

\begin{tcolorbox}[colback=gray!2, colframe=black, boxrule=0.5pt, arc=2pt, width=0.95\linewidth, title=Instruction Template]
\texttt{Your available actions are:\\
Left, Down, Right, Up\\
You can make up to 10 actions, separated by the action separator " || "
}
\end{tcolorbox}

% The environment supports multiple rendering modes for state observation:

% \begin{promptbox}{State Rendering Modes}
% \textbf{Rendering Format:}\\
% Grid cells are separated by tabs horizontally\\
% Rows are separated by newlines\\
% Special handling for start position ('S' replaced with 'F')\\
% Player position updated dynamically in rendering
% \end{promptbox}

\begin{figure}[t]
\centering
\small
\begin{tcolorbox}[colback=gray!2, colframe=black, boxrule=0.5pt, arc=2pt, width=0.95\linewidth, title=Sokoban – Spurious Reasoning Case (3 Turns)]
\textbf{Turn 1:} \\
\textbf{State:}
\begin{verbatim}
######
#___O#
#__X_#
###P##
######
######
\end{verbatim}
\textbf{Output:}  
\texttt{<think>}I've reached the target on the right. I need to move right and push the target.\texttt{</think>}  
\texttt{<answer>}Right || Right || Up\texttt{</answer>}  
Reward: -0.3

\vspace{4pt}
\textbf{Turn 2:} \\
\textbf{State:}
\begin{verbatim}
######
#__XO#
#__P_#
###_##  
######  
######  
\end{verbatim}
\textbf{Output:}  
\texttt{<think>}Now I need to move left and push the box on the left.\texttt{</think>}  
\texttt{<answer>}Left || Left || Up\texttt{</answer>}  
Reward: -0.3

\vspace{4pt}
\textbf{Turn 3:} \\
\textbf{State:}
\begin{verbatim}
######
#P_XO#
#____#
###_##  
######  
######  
\end{verbatim}
\textbf{Output:}  
\texttt{<think>}Now I need to move right and push the box on the right.\texttt{</think>}  
\texttt{<answer>}Right || Right || Up\texttt{</answer>}  
Reward: +1.0
\end{tcolorbox}

\vspace{4pt}
\caption{\textbf{A rollout with spurious reasoning.} The final outcome is successful, but the reasoning traces across turns are inconsistent and sometimes factually incorrect. This reflects a common failure mode where the model optimizes for final rewards while bypassing coherent reasoning, resulting in noisy and potentially misleading supervision during RL training.}
\label{fig:sokoban-spurious}
\vspace{-10pt}
\end{figure}

\section{Generalization Evaluation Environments}
\label{app:generalization-task}

To evaluate generalization beyond the training distribution, we design two new test environments besides the three training environments that vary along different axes:

\begin{itemize}
    \item \textbf{SokobanDifferentGridVocab} modifies the visual vocabulary used to represent the grid. Instead of using the standard symbols (\#, \_, O, X, etc.), it maps grid cells to a new vocabulary such as \texttt{W}, \texttt{G}, \texttt{C}, etc. This tests whether the model generalizes across symbol variations while retaining underlying spatial semantics.
    
    \item \textbf{LargerSokoban} increases the grid size from $6 \times 6$ to $8 \times 8$ and the number of boxes from 1 to 2, introducing greater spatial complexity and longer-horizon planning demands. This setting evaluates whether the policy trained on small puzzles can scale up to more complex configurations.
\end{itemize}

These environments are not seen during training and serve to probe the agent's generalization capability under symbol shift, size scaling, and environment shift, respectively.

\section{Case Study: Spurious Reward from Incorrect Reasoning}
\label{app:case-study-spurious}

While evaluating reasoning behavior in Sokoban, we observe that the model may occasionally receive non-negative or even high rewards despite exhibiting flawed or misleading reasoning traces. Figure~\ref{fig:sokoban-spurious} presents a 3-turn rollout where the model successfully pushes the box onto the target, yet its intermediate decisions reflect incorrect assumptions about the game dynamics.

In Turn 1 and 2, the agent provides plausible but ultimately incoherent plans, such as "pushing the target" or “moving toward the box on the left,” which are either redundant or directionally incorrect. Despite these errors, the final action sequence still manages to reach the goal. Such cases increase reward signal noise, making it harder for RL training to distinguish between genuinely useful plans and coincidentally effective ones.

This highlights a key challenge in multi-turn RL with reasoning agents: \textit{outcome-based reward alone may not sufficiently penalize poor reasoning traces}, especially in environments with sparse or delayed feedback.

\section{Extended Case Studies}

\begin{figure}[t]
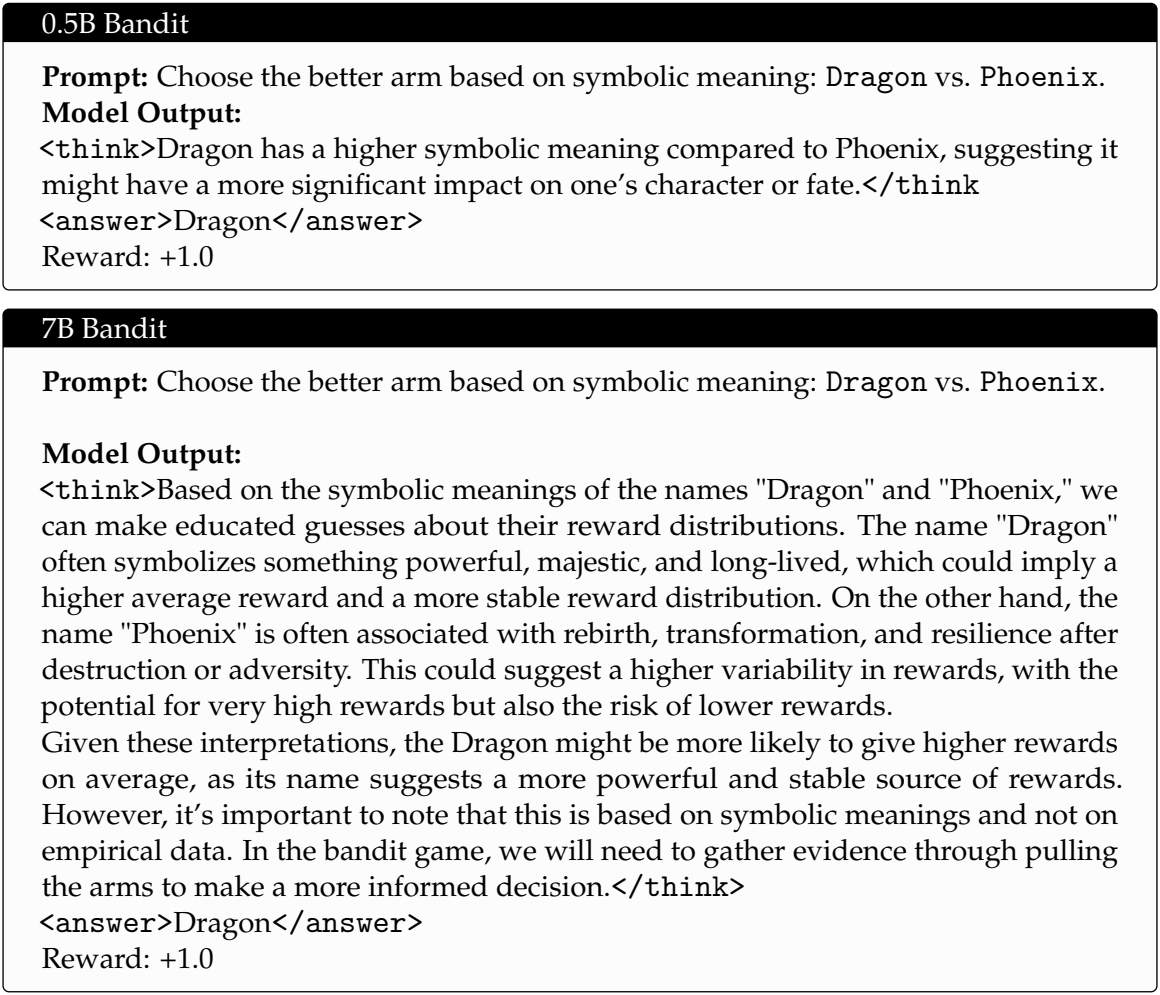

\centering
\begin{tcolorbox}[colback=gray!2, colframe=black, boxrule=0.5pt, arc=2pt, width=0.95\linewidth, title=0.5B Bandit]
\textbf{Prompt:} Choose the better arm based on symbolic meaning: \texttt{Dragon} vs. \texttt{Phoenix}.

\textbf{Model Output:} \\
\texttt{<think>}Dragon has a higher symbolic meaning compared to Phoenix, suggesting it might have a more significant impact on one's character or fate.\texttt{</think} \\
\texttt{<answer>}Dragon\texttt{</answer>} \\
Reward: +1.0
\end{tcolorbox}

\begin{tcolorbox}[colback=gray!2, colframe=black, boxrule=0.5pt, arc=2pt, width=0.95\linewidth, title=7B Bandit]
\textbf{Prompt:} Choose the better arm based on symbolic meaning: \texttt{Dragon} vs. \texttt{Phoenix}.\\

\textbf{Model Output:} \\
\texttt{<think>}Based on the symbolic meanings of the names "Dragon" and "Phoenix," we can make educated guesses about their reward distributions. The name "Dragon" often symbolizes something powerful, majestic, and long-lived, which could imply a higher average reward and a more stable reward distribution. On the other hand, the name "Phoenix" is often associated with rebirth, transformation, and resilience after destruction or adversity. This could suggest a higher variability in rewards, with the potential for very high rewards but also the risk of lower rewards.

Given these interpretations, the Dragon might be more likely to give higher rewards on average, as its name suggests a more powerful and stable source of rewards. However, it's important to note that this is based on symbolic meanings and not on empirical data. In the bandit game, we will need to gather evidence through pulling the arms to make a more informed decision.\texttt{</think>} \\
\texttt{<answer>}Dragon\texttt{</answer>} \\
Reward: +1.0
\end{tcolorbox}
\caption{\textbf{Reasoning-based arm selection in Bandit tasks across model scales.} 
In both cases, the model must infer reward tendencies of symbolic arms (\texttt{Dragon} vs. \texttt{Phoenix}) based on prior knowledge. The 0.5B model offers a short justification rooted in symbolic association. The 7B model generates a more elaborate reasoning chain comparing stability and variance, reflecting its stronger prior knowledge and interpretive capacity. Both ultimately select \texttt{Dragon}, but through different levels of reasoning depth.}
\label{fig:bandit-multiscale}
\end{figure}

To better understand how reasoning quality evolves across model scales and environments, we present representative rollout cases across six settings: Bandit (Figure~\ref{fig:bandit-multiscale}), Sokoban (Figure~\ref{fig:sokoban-multiscale}), FrozenLake (Figure~\ref{fig:frozenlake-multiscale}), and WebShop (Figure~\ref{fig:webshop-multiscale}), each at 0.5B and 7B model scales. We observe that \textbf{larger models tend to generate longer and more coherent reasoning chains, especially in semantic-rich decision tasks like Bandit and WebShop}. However, in grid-based environments like Sokoban and stochastic ones like FrozenLake, \textbf{both small and large models struggle with planning and alignment}, often resorting to brittle heuristics or spurious correlations. These cases align with the experiments in Section~\ref{app:largermodelabaltion} that Sokoban and Frozen Lake does not have a significant performance gain from larger model scales unlike Bandit and WebShop, illustrating how reasoning quality interacts with environment structure, and highlight the challenges of stabilizing reward-grounded reasoning in stochastic or under-specified settings.

\begin{figure}[t]
\centering
\small
\begin{tcolorbox}[colback=gray!2, colframe=black, boxrule=0.5pt, arc=2pt, width=0.95\linewidth, title=0.5B Sokoban – Multi-Turn Rollout]
\textbf{Turn 1:} \\
\textbf{State:}
\begin{verbatim}
######
#__###
#O####
#__XP#
#____#
######
\end{verbatim}
\textbf{Output:} \\
\texttt{<think>}Pushing all boxes to targets.\texttt{</think>} \\
\texttt{<answer>}Left || Left || Down\texttt{</answer>} \\
Reward: -0.3

\vspace{4pt}
\textbf{Turn 2:} \\
\textbf{State:}
\begin{verbatim}
######
#__###
#O####
#X___#
#_P__#
######
\end{verbatim}
\textbf{Output:} \\
\texttt{<think>}Clearing a wall to make progress.\texttt{</think>} \\
\texttt{<answer>}Right || Down || Left\texttt{</answer>} \\
Reward: -0.3

\vspace{4pt}
\textbf{Turn 3:} \\
\textbf{State:}
\begin{verbatim}
######
#__###
#O####
#X___#
#_P__#
######
\end{verbatim}
\textbf{Output:} \\
\texttt{<think>}Pushing another box.\texttt{</think>} \\
\texttt{<answer>}Left || Down || Up\texttt{</answer>} \\
Reward: -0.3
\end{tcolorbox}
\vspace{4pt}
\end{figure}

\begin{figure}[t]
\begin{tcolorbox}[colback=gray!2, colframe=black, boxrule=0.5pt, arc=2pt, width=0.95\linewidth, title=7B Sokoban – Multi-Turn Rollout]
\textbf{Turn 1:} \\
\textbf{State:}
\begin{verbatim}
######
##_P_#
##_X_#
##__O#
###__#
######
\end{verbatim}
\textbf{Output:} \\
\texttt{<think>}First, move the box to target.\texttt{</think>} \\
\texttt{<answer>}Up || Right || Up\texttt{</answer>} \\
Reward: -0.3

\vspace{4pt}
\textbf{Turn 2:} \\
\textbf{State:}
\begin{verbatim}
######
##__P#
##_X_#
##__O#
###__#
######
\end{verbatim}
\textbf{Output:} \\
\texttt{<think>}Now move second box to target.\texttt{</think>} \\
\texttt{<answer>}Left || Down || Left || Down\texttt{</answer>} \\
Reward: -0.4

\vspace{4pt}
\textbf{Turn 3:} \\
\textbf{State:}
\begin{verbatim}
######
##___#
##___#
##PXO#
###__#
######
\end{verbatim}
\textbf{Output:} \\
\texttt{<think>}Finally, move last box to target.\texttt{</think>} \\
\texttt{<answer>}Right || Up || Right || Up\texttt{</answer>} \\
Reward: +1.0
\end{tcolorbox}
\caption{\textbf{Sokoban rollouts across model scales.} At 0.5B, the model performs minimal reasoning and often issues locally valid but suboptimal actions. The 7B model demonstrates more structured planning and symbolic alignment across turns, though still exhibits inefficiencies and heuristic moves in long-horizon settings.}
\label{fig:sokoban-multiscale}
\vspace{-10pt}
\end{figure}

\begin{figure}[t]
\centering
\small
\begin{tcolorbox}[colback=gray!2, colframe=black, boxrule=0.5pt, arc=2pt, width=0.95\linewidth, title=0.5B FrozenLake – Multi-Turn Rollout]
\textbf{State:}
\begin{verbatim}
_P_G
____
O_OO
O___
\end{verbatim}

\textbf{Model Output \textcolor{red}{(repeated each turn)}:} \\
\texttt{<think>}Forbid the hole and go to the target, I should go left then go up.\texttt{</think>} \\
\texttt{<answer>}Left || Up\texttt{</answer>} \\
Reward: 0.0 (repeated for 4 turns)

\vspace{4pt}
\textbf{Final Turn (Turn 5):} \\
\textbf{State:}
\begin{verbatim}
P__G
____
O_OO
O___
\end{verbatim} \\
\textbf{Model Output:} \\
\texttt{<think>}Forbid the hole and go to the target, I should go left then go up.\texttt{</think>} \\
\texttt{<answer>}Left || Up\texttt{</answer>} \\
Reward: 0.0
\end{tcolorbox}
\vspace{4pt}
\begin{tcolorbox}[colback=gray!2, colframe=black, boxrule=0.5pt, arc=2pt, width=0.95\linewidth, title=7B FrozenLake – Multi-Turn Rollout]
\textbf{Turn 1:} \\
\textbf{State:}
\begin{verbatim}
____
_P__
OGO_
____
\end{verbatim} \\
\textbf{Model Output:} \\
\texttt{<think>}Fill holes, then go target.\texttt{</think>} \\
\texttt{<answer>}Left || Up\texttt{</answer>} \\
Reward: +1.0 (due to slipping into goal)

\textit{Note: although the action was incorrect, the agent reached the goal due to stochastic slipping.}
\end{tcolorbox}
\caption{\textbf{FrozenLake rollouts across model scales.} At 0.5B, the agent repeats a fixed plan regardless of outcome, suggesting limited adaptation or planning. The 7B agent receives high reward despite issuing a suboptimal command, due to stochastic transitions—highlighting the difficulty of credit assignment in such environments and the risk of reinforcing spurious patterns.}
\label{fig:frozenlake-multiscale}
\vspace{-10pt}
\end{figure}

\begin{figure}[t]
\begin{tcolorbox}[colback=gray!2, colframe=black, boxrule=0.5pt, arc=2pt, width=0.95\linewidth, title=0.5B WebShop – Multi-Turn Rollout]
\textbf{Turn 1:} \\
\textbf{State:}
Instruction: Find me super soft decorative pillows for living room with color: beige latte, and size: 20''x20'', and price lower than 30.00 dollars.We must buy a product within 10 actions. It doesn't have to match perfectly with description.
Search term should not include details like size, ...\\
\textbf{Model Output:} \\
\texttt{<think>}To find a super soft decorative pillows for the living room with a beige latte color and a size of 20''x20'', I will start with a search for pillows with a beige color and a size of 20''x20''.\texttt{</think>} \\
\texttt{<answer>}search[bisque pillow with beige color 20''x20']\texttt{</answer>} \\
Reward: 0.0

\vspace{4pt}
\textbf{Turn 2:} \\
\textbf{State:}
Instruction: [SEP] Find me super soft decorative pillows for living room with color: beige latte, and size: 20''x20'', and price lower than 30.00 dollars [SEP] Back to Search [SEP] Page 1 (Total results: 50) [SEP] Next \texttt{>} [SEP] B08L2ZDWN2 [SEP] PAVILIA Decorative Sherpa Throw Pillow Covers, ... click[b08qvdnjc7], click[b07p24zmxk], click[b07n33bpq6], click[b09nrf2qgd],\\
\textbf{Model Output:} \\
\texttt{<think>}Searching for pillows with beige color and 20''x20'' size, and price lower than \$30.\texttt{</think>} \\
\texttt{<answer>}click[b08l2zdwn2]\texttt{</answer>} \\
Reward: 0.0

\vspace{4pt}
\textbf{Turn 3:} \\
\textbf{State:}
Instruction: [SEP] Find me super soft decorative pillows for living room with color: beige latte, and size: 20''x20'', and price lower than 30.00 dollars [SEP] Back to Search [SEP] \texttt{<} Prev [SEP] color [SEP] beige latte [SEP] black [SEP] blush pink [SEP] dusty blue [SEP] emerald green [SEP] ... grey], click[mustard yellow], click[teal blue], click[white], click[18''x18''],\\
\textbf{Model Output:} \\
\texttt{<think>}Searching for pillows with beige color and 20''x20'' size, and price lower than \$30.\texttt{</think>} \\
\texttt{<answer>}click[beige latte]\texttt{</answer>} \\
Reward: 0.0
...

\textbf{Turn 9:} \\
\textbf{State:}
Instruction: [SEP] Find me super soft decorative pillows for living room with color: beige latte, and size: 20''x20'', and price lower than 30.00 dollars [SEP] Back to Search [SEP] \texttt{<} Prev [SEP] color [SEP] beige latte [SEP] black [SEP] blush pink [SEP] dusty blue [SEP] emerald green [SEP] ... grey], click[mustard yellow], click[teal blue], click[white], click[18''x18''],\\
\textbf{Model Output:} \\
\texttt{<think>}Searching for pillows with beige color and 20''x20'' size, and price lower than \$30.\texttt{</think>} \\
\texttt{<answer>}click[black]\texttt{</answer>} \\
Reward: 0.0

\end{tcolorbox}
\end{figure}

\begin{figure}[t]
\small
\begin{tcolorbox}[colback=gray!2, colframe=black, boxrule=0.5pt, arc=2pt, width=0.95\linewidth, title=3B WebShop – Multi-Turn Rollout]
\textbf{Turn 1:} \\
\textbf{State:}
Instruction: Find me wall mounted, white item, easy clean tv stands for living room with color: white+brown, and price lower than 130.00 dollars.We must buy a product within 10 actions. It doesn't have to match perfectly with description.
Search term should not include details like size, ... \\
\textbf{Model Output:} \\
\texttt{<think>}Let's start by searching for the items within the given criteria and narrow down the options. The search term will be 'white+brown tv stand living room' to find wall-mounted, easy-clean TV stands with the specified colors and in the living room category.\texttt{</think>} \\
\texttt{<answer>}search[white+brown tv stand living room]\texttt{</answer>} \\
Reward: 0.0

\vspace{4pt}
\textbf{Turn 2:} \\
\textbf{State:}
Instruction: [SEP] Find me wall mounted, white item, easy clean tv stands for living room with color: white+brown, and price lower than 130.00 dollars [SEP] Back to Search [SEP] Page 1 (Total results: 50) [SEP] Next \texttt{>} [SEP] B09F6SCW47 [SEP] hongxinq Floating TV Stand with Storage Shelves, ... click[b079n3vlrj], click[b0924pklq4], click[b09lhm4wqs], click[b08k7ldm7q],\\
\textbf{Model Output:} \\
\texttt{<think>}After reviewing the search results, the first item that seems to match the criteria is the 'Floating TV Stand with Storage Shelves, Chipboard Wall Mounted Hanging Television Cabinet' from 'B09F6SCW47'. It is white with brown accents, fits the living room and bedroom category, and is priced at \$90.79, which is lower than the \$130.00 limit. It's also the closest match in terms of color and style. Since it's the first matching option, I will prioritize clicking on it to move forward with the purchase. However, I'll make sure to click [size] and then [color] if they are available and required to finalize the purchase.\texttt{</think>} \\
\texttt{<answer>}click[b09f6scw47]\texttt{</answer>} \\
Reward: 0.0

\vspace{4pt}

...

\textbf{Turn 7:} \\
\textbf{State:}
Instruction: [SEP] Find me wall mounted, white item, easy clean tv stands for living room with color: white+brown, and price lower than 130.00 dollars [SEP] Back to Search [SEP] \texttt{<} Prev [SEP] color [SEP] brown [SEP] concrete gray [SEP] grey [SEP] white [SEP] white+brown [SEP] hongxinq ... click[buy now], click[brown], click[concrete gray], click[grey], click[white],\\
\textbf{Model Output:} \\
\texttt{<think>}With the 'Floating TV Stand with Storage Shelves, Chipboard Wall Mounted Hanging Television Cabinet' from 'B09F6SCW47' confirmed to have the correct white and brown color options and the required size of 12Inx11.8Inx35.4In, I will now proceed to click the 'buy now' button to finalize the purchase. Given that we have 3 actions left and the product matches the criteria closely, I will click on 'buy now' to complete the transaction. This ensures that we stay within the action limit and make a purchase that closely aligns with the specified requirements.\texttt{</think>} \\
\texttt{<answer>}click[buy now]\texttt{</answer>} \\
Reward: 1.0
\end{tcolorbox}
\caption{\textbf{WebShop rollouts illustrate the impact of model scale on long-context decision making.} 
At 0.5B, the agent becomes stuck in a loop, repeatedly selecting irrelevant options despite rich contextual information—
indicating difficulty in long-horizon memory and goal tracking. In contrast, the 3B model executes a successful multi-step 
reasoning chain: narrowing search queries, navigating product options, selecting attributes, and finalizing purchase. 
This highlights the importance of scale for compositional planning in realistic, open-domain environments.}
\label{fig:webshop-multiscale}
\end{figure}

\clearpage

\end{CJK*}
\end{document}